\pdfoutput=1

\documentclass[11pt]{article}

\usepackage{ACL2023}

\usepackage{times}
\usepackage{latexsym}
\usepackage{CJKutf8}
\usepackage{amsmath}
\usepackage{verbatim}
\usepackage{booktabs}
\usepackage{amssymb}
\usepackage{pifont}
\usepackage{stfloats}
\usepackage{multicol}
\usepackage{float}
\usepackage{graphicx}
\usepackage{subfigure}
\usepackage{multirow}
\usepackage{array}
\newcommand{\PreserveBackslash}[1]{\let\temp=\\#1\let\\=\temp}
\newcommand{\bm}[1]{\mbox{\boldmath{$#1$}}}
\newcolumntype{C}[1]{>{\PreserveBackslash\centering}p{#1}}
\newcolumntype{R}[1]{>{\PreserveBackslash\raggedleft}p{#1}}
\newcolumntype{L}[1]{>{\PreserveBackslash\raggedright}p{#1}}
\usepackage{diagbox}
\usepackage[noend]{algpseudocode}
\usepackage{booktabs}
\usepackage{amsfonts}
\usepackage{bbm}
\usepackage{mathtools}

\usepackage{xcolor}
\usepackage[linesnumbered,ruled,vlined]{algorithm2e}

\SetCommentSty{mycommfont}

\SetKwInput{KwInput}{Input}                
\SetKwInput{KwOutput}{Output}              
\SetKwInput{KwInit}{Initialization}

\usepackage[T1]{fontenc}

\usepackage[utf8]{inputenc}

\usepackage{microtype}

\usepackage{inconsolata}

\title{End-to-End Simultaneous Speech Translation with \\Differentiable Segmentation}

\author{Shaolei Zhang \textsuperscript{\rm 1,2},
    Yang Feng \textsuperscript{\rm 1,2}\thanks{ \ \ Corresponding author: Yang Feng.} \\
        \textsuperscript{\rm 1}{Key Laboratory of Intelligent Information Processing} \\ Institute of Computing Technology, Chinese Academy of Sciences (ICT/CAS) \\
    { \textsuperscript{\rm 2} {University of Chinese Academy of Sciences, Beijing, China}} \\
     \texttt{\{\href{mailto:zhangshaolei20z@ict.ac.cn}{zhangshaolei20z}, \href{mailto:fengyang@ict.ac.cn}{fengyang}\}@ict.ac.cn}  }

\begin{document}
\maketitle
\begin{abstract}
End-to-end simultaneous speech translation (SimulST) outputs translation while receiving the streaming speech inputs (a.k.a. streaming speech translation), and hence needs to segment the speech inputs and then translate based on the current received speech. However, segmenting the speech inputs at unfavorable moments can disrupt the acoustic integrity and adversely affect the performance of the translation model. Therefore, learning to segment the speech inputs at those moments that are beneficial for the translation model to produce high-quality translation is the key to SimulST. Existing SimulST methods, either using the fixed-length segmentation or external segmentation model, always separate segmentation from the underlying translation model, where the gap results in segmentation outcomes that are not necessarily beneficial for the translation process. In this paper, we propose \emph{Differentiable Segmentation} (\emph{DiSeg}) for SimulST to directly learn segmentation from the underlying translation model. DiSeg turns hard segmentation into differentiable through the proposed expectation training, enabling it to be jointly trained with the translation model and thereby learn translation-beneficial segmentation. Experimental results demonstrate that DiSeg achieves state-of-the-art performance and exhibits superior segmentation capability\footnote{$\:$Code is available at \url{https://github.com/ictnlp/DiSeg}.}.

\end{abstract}

\section{Introduction}

End-to-end simultaneous speech translation (SimulST) \citep{10.2307/30219116,oda-etal-2014-optimizing,ren-etal-2020-simulspeech,zeng-etal-2021-realtrans,zhang-etal-2022-learning} outputs translation when receiving the streaming speech inputs, and is widely used in real-time scenarios such as international conferences, live broadcasts and real-time subtitles. Compared with the offline speech translation waiting for the complete speech inputs \citep{Weiss2017,wang-etal-2020-fairseq}, SimulST needs to segment the streaming speech inputs and synchronously translate based on the current received speech, aiming to achieve high translation quality under low latency \citep{hamon-etal-2009-end,Cho2016,ma-etal-2020-simulmt,dualpath}.

However, it is non-trivial to segment the streaming speech inputs as the speech always lacks explicit boundary \citep{zeng-etal-2021-realtrans}, and segmenting at unfavorable moments will break the acoustic integrity and thereby drop the translation performance \citep{dong-etal-2022-learning}. Therefore, the precise segmentation of streaming speech is the core challenge of SimulST task \citep{zhang-etal-2022-learning}. To ensure that the speech representations derived from the segmentation results can produce high-quality translation, SimulST model should learn a translation-beneficial segmentation from the underlying translation model.

\begin{figure}[t]
\centering
\subfigure[Fixed segmentation with the equal length of 280$ms$.]{
\includegraphics[width=3in]{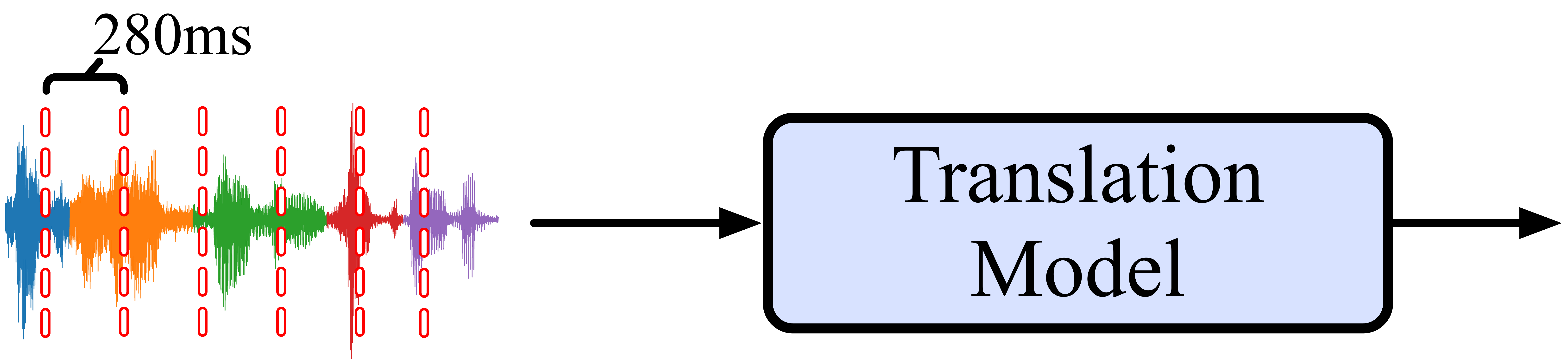}\label{fig:ill1}
}
\subfigure[Adaptive segmentation with external segmentation model.]{
\includegraphics[width=3in]{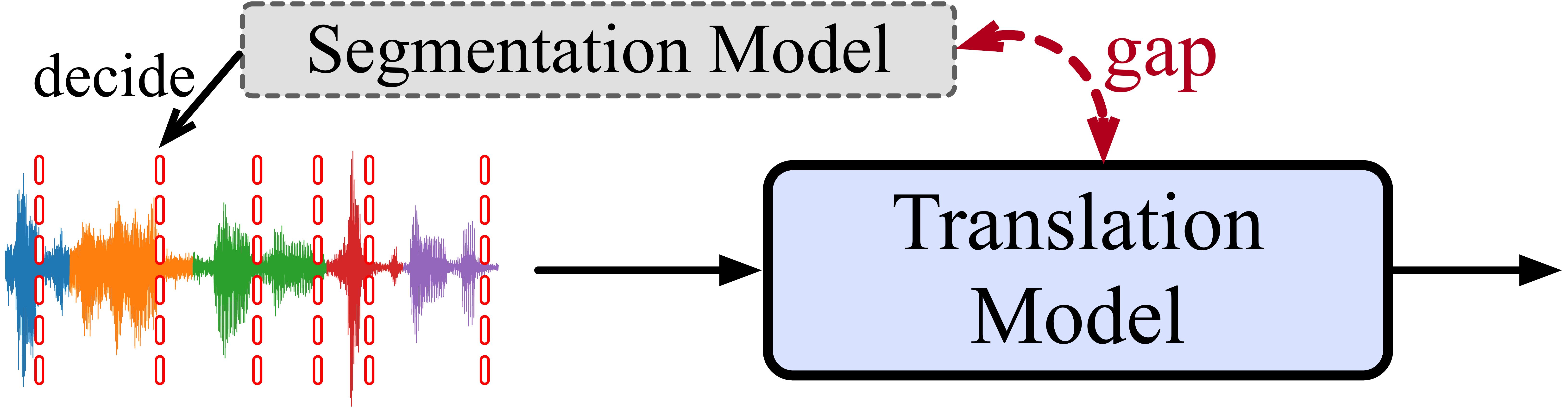}\label{fig:ill2}
}
\subfigure[Differentiable segmentation within translation model.]{
\includegraphics[width=3in]{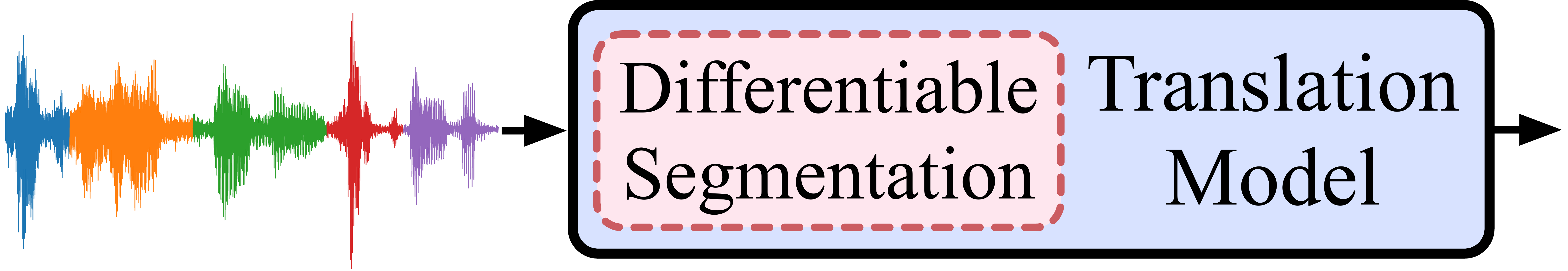}\label{fig:ill3}
}
\caption{Illustration of differentiable segmentation (DiSeg) compared with the previous methods.}
\label{fig:ill}
\end{figure}

Existing SimulST methods, involving fixed and adaptive, always fail to learn the segmentation directly from the underlying translation model. The fixed method divides the streaming inputs based on the equal length, e.g., 280$ms$ per segment \citep{ma-etal-2020-simulmt,9414897,9414276}, as shown in Figure \ref{fig:ill1}. Such methods completely ignore the translation model and always break the acoustic integrity \citep{dong-etal-2022-learning}. The adaptive method dynamically decides the segmentation and thereby achieves better SimulST performance, as shown in Figure \ref{fig:ill2}. However, previous adaptive methods often use the external segmentation model \citep{ma-etal-2020-simulmt, zhang-etal-2022-learning} or heuristic detector \citep{zeng-etal-2021-realtrans,chen-etal-2021-direct,dong-etal-2022-learning} for segmentation, which leave a gap between the segmentation and translation model. This gap hinders learning segmentation directly from the translation model \citep{Arivazhagan2019}, hence making them difficult to get segmentation results that are most beneficial to translation quality.

Under these grounds, we aim to integrate the segmentation into translation model, and thereby directly learn segmentation from the underlying translation model, as shown in Figure \ref{fig:ill3}. To this end, we propose \emph{Differentiable Segmentation} (\emph{DiSeg}) for SimulST, which can be jointly trained with the underlying translation model. DiSeg employs a Bernoulli variable to indicate whether the streaming speech inputs should be segmented or not. Then, to address the issue that hard segmentation precludes back-propagation (i.e., learning) from the underlying translation model, we propose an expectation training to turn the segmentation into differentiable. Owing to powerful segmentation, DiSeg can handle simultaneous and offline speech translation through a unified model. Experiments show that DiSeg achieves state-of-the-art performance on SimulST, while also delivering high-quality offline speech translation.

\section{Background}
\label{sec:Background}
\textbf{Offline Speech Translation} The corpus of speech translation task is always denoted as the triplet $\mathcal{D}\!=\!\left\{\left(\mathbf{s},\mathbf{x},\mathbf{y} \right)\right\}$, where $\mathbf{s}\!=\!\left ( s_{1},\cdots ,s_{\left|\mathbf{s} \right|}  \right )$ is source speech, $\mathbf{x}\!=\!\left ( x_{1},\cdots ,x_{\left|\mathbf{x} \right|}  \right )$ is source transcription and $\mathbf{y}\!=\!\left ( y_{1},\cdots ,y_{\left|\mathbf{y} \right|}  \right )$ is target translation. The mainstream speech translation architecture often consists of an acoustic feature extractor and a translation model following \citep{nguyen2020investigating}. Acoustic feature extractor extracts speech features $\mathbf{a}\!=\!\left ( a_{1},\cdots ,a_{\left|\mathbf{a} \right|}  \right )$ from source speech $\mathbf{s}$, which is often realized by a pre-trained acoustic model \citep{NEURIPS2020_92d1e1eb}. Then, the translation model, realized by a Transformer model \citep{NIPS2017_7181}, generates $\mathbf{y}$ based on all speech features $\mathbf{a}$. During training, existing methods always improve speech translation performance through multi-task learning \cite{anastasopoulos-chiang-2018-tied,tang-etal-2021-improving,9415058}, including speech translation, automatic speech recognition (ASR) and machine translation (MT) (add a word embedding layer for MT task), where the learning objective $\mathcal{L}_{mtl}$ is:
\begin{gather}
    \mathcal{L}_{mtl}=\mathcal{L}_{st}+\mathcal{L}_{asr}+\mathcal{L}_{mt},
\end{gather}
where $\mathcal{L}_{st}$, $\mathcal{L}_{asr}$ and $\mathcal{L}_{mt}$ are the cross-entropy loss of pairs $\mathbf{s}\!\rightarrow\!\mathbf{y}$, $\mathbf{s}\!\rightarrow\!\mathbf{x}$ and $\mathbf{x}\!\rightarrow\!\mathbf{y}$, respectively.

\textbf{Simultaneous Translation} (SimulST) Unlike offline speech translation, SimulST needs to decide when to segment the inputs and then translate based on the received speech features \citep{ren-etal-2020-simulspeech,ma-etal-2020-simulmt}. Since the decisions are often made based on the speech features after downsampling, we use $g\!\left ( t \right )$ to denote the number of speech features when the SimulST model translates $y_{t}$, where speech features $\mathbf{a}_{\leq g\left ( t \right )}$ are extracted from the current received speech $\mathbf{\hat{s}}$. Then, the probability of generating $y_{t}$ is $p\left ( y_{t}\mid \mathbf{\hat{s}},\mathbf{y}_{< t} \right )$. How to decide $g\!\left ( t \right )$ is the key of SimulST, which should be beneficial for the translation model to produce the high-quality translation.

\section{Method}
\label{sec:method}
\begin{figure}[t]
    \centering
    \includegraphics[width=3.03in]{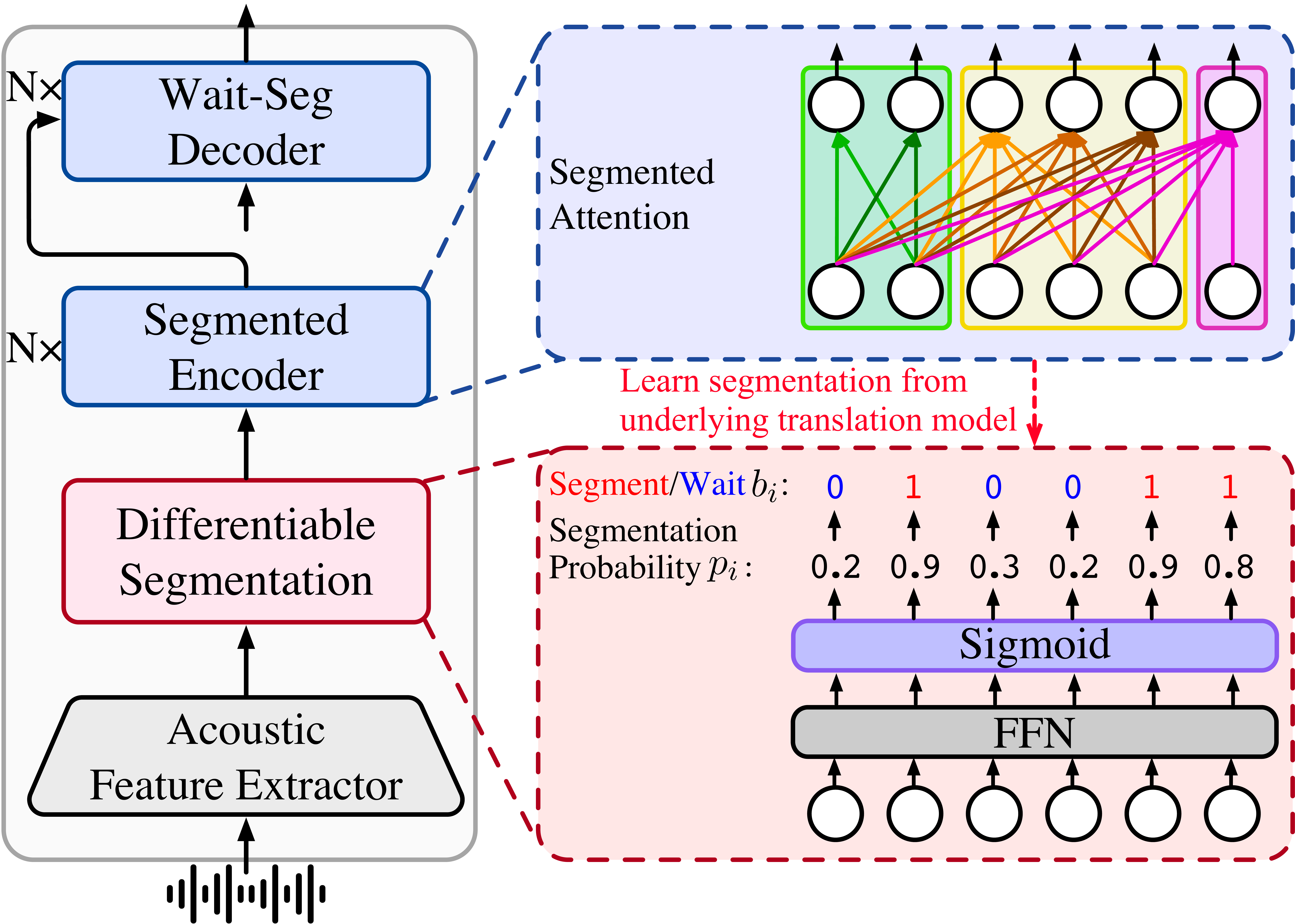}
    \caption{Architecture of the proposed DiSeg.}
    \label{fig:model}
\end{figure}

We propose differentiable segmentation (DiSeg) to learn segmentation directly from the translation model, aiming to achieve translation-beneficial segmentation. As shown in Figure \ref{fig:model}, DiSeg predicts a Bernoulli variable to indicate whether to segment, and then turns the hard segmentation into differentiable through the proposed expectation training, thereby jointly training segmentation with the translation model. We will introduce the segmentation, training and inference of DiSeg following.

\subsection{Differentiable Segmentation}
\label{sec:Segmentation}

To segment the streaming speech inputs, DiSeg predicts a Bernoulli variable $0/1$ for each speech feature, corresponding to waiting or segmenting. Specifically, for the speech feature $a_{i}$, DiSeg predicts a Bernoulli segmentation probability $p_{i}$, corresponding to the probability of segmenting the speech at $a_{i}$. Segmentation probability $p_{i}$ is calculated through a feed-forward network (FFN) followed by a sigmoid activation, and then $p_{i}$ is used to parameterize the Bernoulli variable $b_{i}$:
\begin{align}
p_{i}=&\;\mathrm{Sigmoid}\!\left ( \mathrm{FFN}\!\left ( a_{i} \right ) \right ), \label{eq:energy}\\
b_{i} \sim &\;\mathrm{Bernoulli} \!\left ( p_{i} \right ).
\end{align}
If $b_{i}\!=\!1$, DiSeg segments the streaming speech at $a_{i}$; If $b_{i}\!=\!0$, DiSeg waits for more inputs. In inference, DiSeg sets $b_{i}\!=\!1$ if $p_{i}\!\geq \!0.5$, and sets $b_{i}\!=\!0$ if $p_{i}\!<\! 0.5$ \citep{LinearTime}.

\textbf{Segmented Attention} After segmenting the speech features, we propose segmented attention for the encoder of the translation model, which is an attention mechanism between uni-directional and bi-directional attention. In segmented attention, each speech feature can focus on the features in the same segment and the previous segments (i.e., bi-directional attention within a segment, uni-directional attention between segments), as shown in Figure \ref{fig:attn1}. In this way, segmented attention can not only satisfy the requirement of encoding streaming inputs in SimulST task (i.e., the characteristic of uni-directional attention) \citep{multipath,zeng-etal-2021-realtrans}, but also capture more comprehensive context representations of segments (i.e., the advantage of bi-directional attention).

\subsection{Expectation Training}
\label{sec:training}

\begin{figure}[t]
\centering
\subfigure[Segmented attention.]{
\includegraphics[width=1.25in]{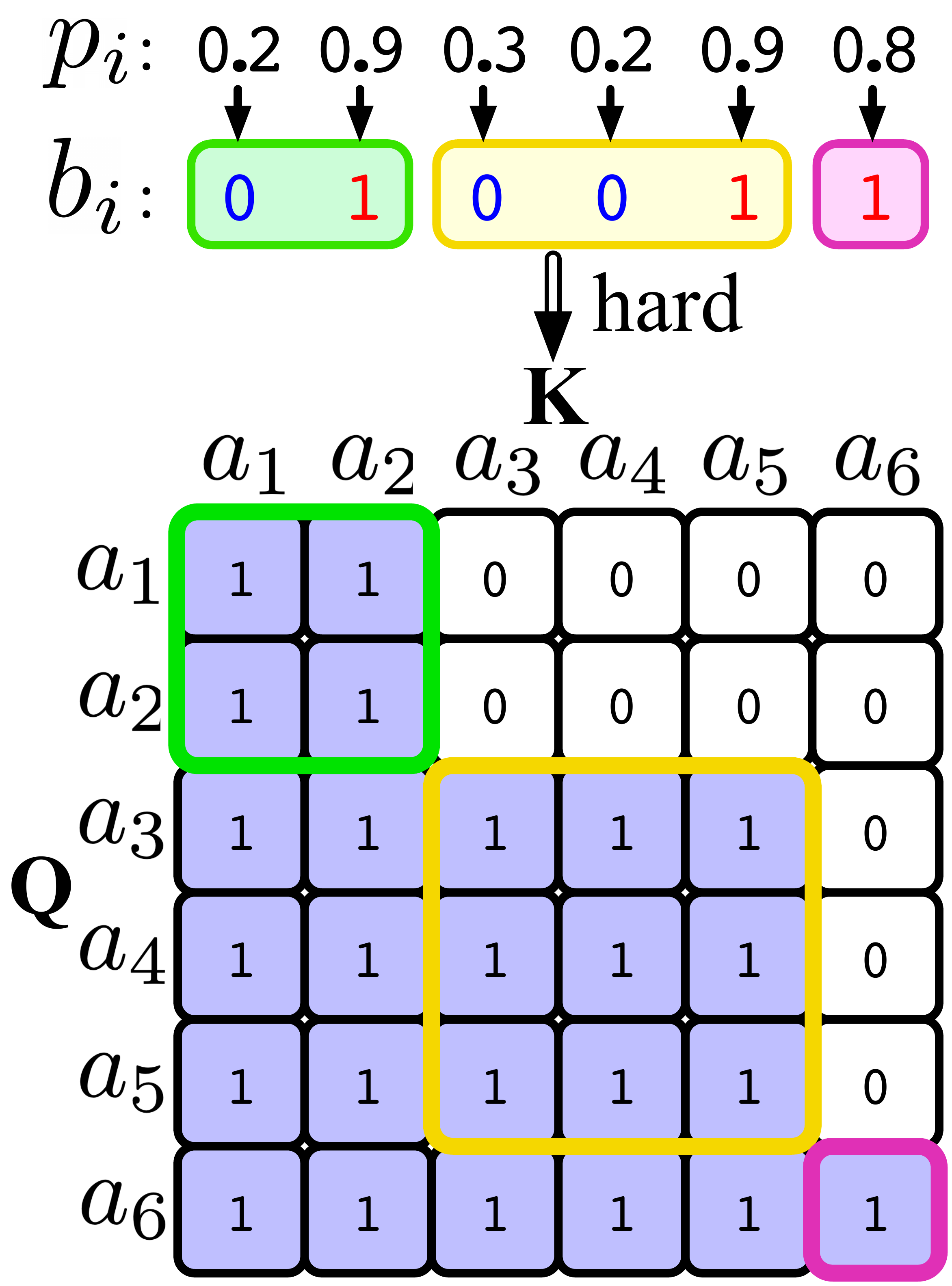}\label{fig:attn1}
}
\subfigure[Expected segmented attention.]{
\includegraphics[width=1.59in]{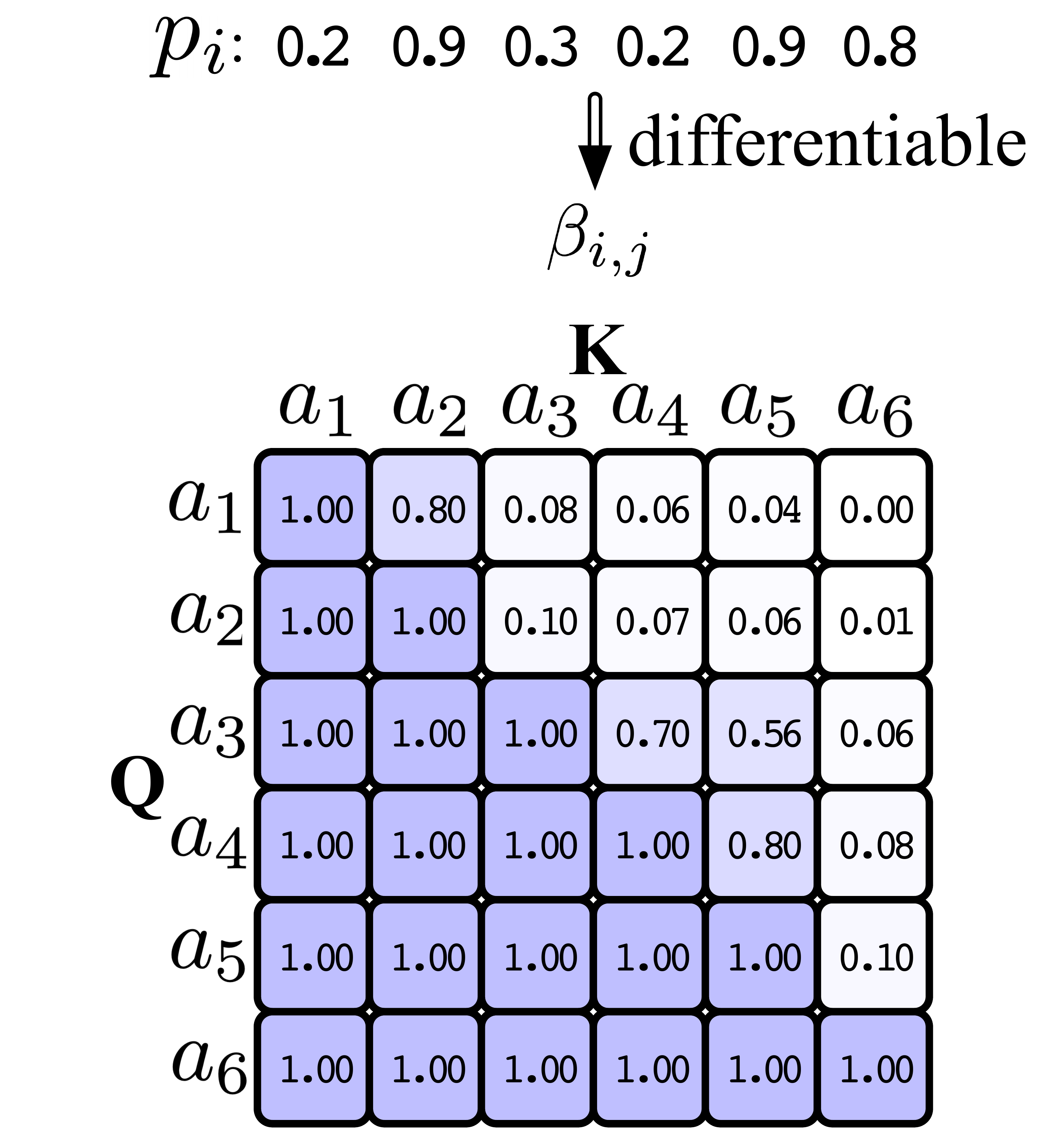}\label{fig:attn2}
}
\caption{Schematic diagram of segmented attention in inference and expected segmented attention in training.}
\label{fig:attn}
\end{figure}
Hard segmentation based on Bernoulli variable $b_{i}$ precludes back-propagation (i.e., learning) from the translation model to the segmentation probability $p_{i}$ during training. To address this, we propose expectation training to turn segmentation into differentiable. In expectation training, we first constrain the number of segments, and then learns segmentation from the translation model at both acoustic and semantic levels, which are all trained in expectation via the segmentation probability $p_{i}$.

\textbf{Learning Segment Number} To avoid too many segments breaking the acoustic integrity or too few segments degenerating the model into offline translation, we need to constrain the total number of segments. Intuitively, the source speech should be divided into $K$ segments, where $K$ is the number of words in the source transcription, so that each speech segment can correspond to a complete word. To this end, we apply $\mathcal{L}_{num}$ to constrain the expected segment number to be $K$. In particular, in order to prevent excessive segmentation on consecutive silent frames, we also to encourage only one segmentation in several consecutive speech frames. Therefore, $\mathcal{L}_{num}$ is calculated as:
\begin{align}
    &\mathcal{L}_{num}=\left\|\sum_{i=1}^{\left|\mathbf{a} \right|}p_{i}-K \right\|_{2} +\\ &\;\;\;\;\;\;\;\;\;\left\|\sum \mathrm{MaxPool}\left(p_{i}, \left \lfloor\frac{\left|\mathbf{a} \right|}{K}\right \rfloor\right)-K \right\|_{2},
\end{align}
where $\sum_{i=1}^{\left|\mathbf{a} \right|}p_{i}$ is the expected segment number and $\mathrm{MaxPool}\left(\cdot\right)$ is the max polling operation with kernel size of $\left \lfloor \left|\mathbf{a} \right|/K\right \rfloor$.

To make the effect of $p_{i}$ in expectation training match $b_{i}$ in the inference, we hope that $p_{i}\!\approx \!0$ or $p_{i}\!\approx \!1$ and thereby make $p_{i}\!\approx \!b_{i}$. To achieve this, we aim to encourage the discreteness of segmentation probability $p_{i}$ during training. Following \citet{SALAKHUTDINOV2009969,NIPS2016_c7635bfd,LinearTime}, a straightforward and efficient method is adding a Gaussian noise before the sigmoid activation in Eq.(\ref{eq:energy}). Formally, in expectation training, Eq.(\ref{eq:energy}) is rewritten as:
\begin{gather}
    p_{i}=\mathrm{Sigmoid}\!\left ( \mathrm{FFN}\!\left ( a_{i} \right )+\mathcal{N}\!\left(0,n \right) \right ), \label{eq:noise}
\end{gather}
where $\mathcal{N}\!\left(0,n \right)$ is a Gaussian noise with $0$ as mean and $n$ as variance. Noise is only applied in training.

\textbf{Learning Segmentation at Acoustic Level} A good segmentation should avoid breaking acoustic integrity and benefit the underlying translation model. As mentioned in Sec.\ref{sec:Segmentation}, the encoder of the translation model applies the segmented attention to model the correlation between speech features and get the source representations. Correspondingly, we propose \emph{expected segmented attention} to turn the hard segmentation into differentiable during training, thereby directly learning translation-beneficial segmentation from the translation model.

In segmented attention during inference, speech feature $a_{i}$ can only pay attention to feature $a_{j}$ that locates in the same segment or previous segments, and mask out the rest features. In expected segmented attention, to enable back-propagation, we introduce the probability that $a_{i}$ can pay attention to $a_{j}$ instead of the hard segmentation, denoted as $\beta_{i,j}$. As shown in Figure \ref{fig:attn2}, $\beta_{i,j}$ measures the probability that $a_{j}$ locates in the same segment as $a_{i}$ or in the segment before $a_{i}$, calculated as:
\begin{gather}
    \beta_{i,j}=\begin{cases}
\prod_{l=i}^{j-1}\!\left ( 1-p_{l} \right ), & \text{ if } i<j \\
1\;\;\;\;\;\;\;\;\;\;\;\;\;\;\;\;\;\;\;, & \text{ if } i\geq j 
\end{cases}.\label{eq:expected segmented attention}
\end{gather}
If $a_{j}$ lags behind $a_{i}$, the premise that $a_{i}$ and $a_{j}$ are in the same segment is that no segmentation is between $a_{i}$ and $a_{j-1}$, i.e., $\prod_{l=i}^{j-1}\!\left ( 1-p_{l} \right )$. If $a_{j}$ is before $a_{i}$, $a_{i}$ can necessarily focus on $a_{j}$. Then, $\beta_{i,j}$ is multiplied with the original soft attention $\alpha_{i,j}$, and normalized to get the final attention $\gamma_{i,j}$:
\begin{align}
    \tilde{\gamma}_{i,j} =&\alpha _{i,j}\times \beta _{i,j},\\\gamma_{i,j}=&\tilde{\gamma}_{i,j}/\sum_{l=1}^{\left|\mathbf{a} \right|}\tilde{\gamma}_{i,l}.
\end{align}
Finally, $\gamma_{i,j}$ is used to calculate the context vector. Owing to the expected segmented attention, $p_{i}$ can be jointly trained with the translation model via cross-entropy loss $\mathcal{L}_{mtl}$. Specifically, if the underlying translation model prefers to let $a_{i}$ pay attention to the subsequent $a_{j}$ for better source representation (i.e., a large $\gamma_{i,j}$), the probability $\beta_{i,j}$ that they are in the same segment will be increased, which teaches DiSeg to prevent segmenting between $a_{i}$ and $a_{j}$. In this way, DiSeg can avoid breaking acoustic integrity and learn the segmentation that is beneficial to the translation model.

\begin{figure}[t]
    \centering
    \includegraphics[width=3in]{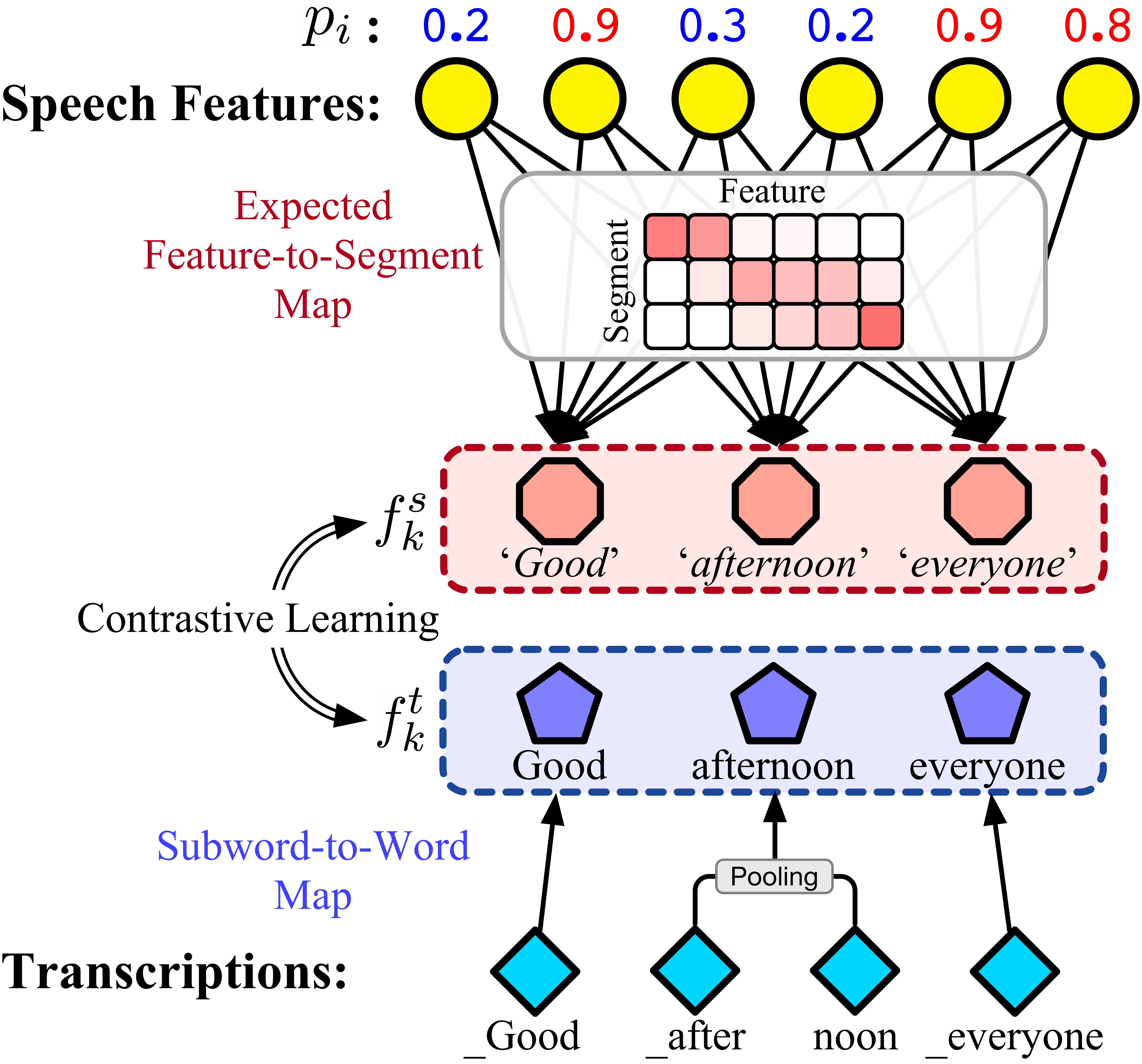}
    \caption{Schematic diagram of learning segmentation at the semantic level, via two mappings.}
    \label{fig:ctr}
\end{figure}

\textbf{Learning Segmentation at Semantic Level} Besides encouraging the related speech features to locate in the same segment via expected segmented attention, we aim to further learn segmentation at the semantic level. In the multi-task learning framework, the transcription $\mathbf{x}$ is monotonically corresponding to the source speech, so transcription is a good choice to provide semantic supervision for segmentation. However, there is a significant gap between transcription representations and speech features in sequence length \citep{liu2020bridging,zeng-etal-2021-realtrans}, so how to align them is key challenge for semantic supervision. 
Fortunately, the proposed differentiable segmentation divides the speech into $K$ segments, where $K$ is also the word number in transcription. Therefore, both speech and transcription sequences can be mapped to the sequence of length $K$ and accordingly reach an agreement on the corresponding representation to achieve semantic supervision, as shown in Figure \ref{fig:ctr}.

For transcription $\mathbf{x}$, since it is a sequence of subwords after tokenization \citep{kudo-richardson-2018-sentencepiece}, we introduce a \emph{subword-to-word map} to get the representation of the whole word. Given the text embedding $\mathbf{e}\!=\!\mathrm{Emb}\!\left(\mathbf{x} \right)$ of subwords, the whole word representation is the average pooling result on the embedding of all subwords it contains. Formally, the representation $f^{t}_{k}$ of the $k^{th}$ word that consists of subwords $\mathbf{x}\left [ l_{k}:r_{k} \right ]$ is calculated as:
\begin{gather}
f^{t}_{k}=\frac{1}{r_{k}-l_{k}+1}\sum_{i=l_{k}}^{r_{k}} e_{i}.
\end{gather}

For speech, the segment representations also need to be differentiable to segmentation (i.e., $p_{i}$), thereby enabling DiSeg to learn segmentation at semantic level. To this end, we propose an \emph{expected feature-to-segment map} to get the expected segment representations. Expected feature-to-segment map does not forcibly assign speech feature $a_{i}$ to a certain segment but calculates the probability that $a_{i}$ belongs to the $k^{th}$ segment, denoted as $p\!\left ( a_{i} \in \mathrm{Seg}_{k} \right )$, which can be calculated via dynamic programming (refer to Appendix \ref{sec:dp} for details):
\begin{gather}
\begin{aligned}
    p\!\left ( a_{i} \in \mathrm{Seg}_{k} \right )= p\!\left (a_{i-1} \in \mathrm{Seg}_{k-1}  \right )\times p_{i-1}& \\
    +\; p\!\left (a_{i-1} \in \mathrm{Seg}_{k}  \right )\times \left (1- p_{i-1} \right ) &.
\end{aligned}
\end{gather}
Then, the expected representation $f^{s}_{k}$ of the $k^{th}$ segment is calculated by weighting all speech features:
\begin{gather}
f^{s}_{k}=\sum_{i=1}^{\left|\mathbf{a} \right|} p\!\left ( a_{i} \in \mathrm{Seg}_{k} \right ) \times a_{i}.
\end{gather}

Owing to the proposed two mappings, transcription and speech are mapped to the representations with the same length, i.e., $K$ segments/words, where $f^{s}_{k}$ corresponds to $f^{t}_{k}$. To provide semantic supervision for segmentation, we apply multi-class N-pair contrastive loss $\mathcal{L}_{ctr}$ \citep{NIPS2016_6b180037} between $\mathbf{f}^{s}$ and $\mathbf{f}^{t}$, where $f^{t}_{k}$ is the positive sample of $f^{s}_{k}$ and the rest are the negative samples, calculated as:
\begin{gather}
    \mathcal{L}_{ctr}\!=\!-\!\!\!\sum _{\mathbf{f}^{s},\mathbf{f}^{t}} \!\log\frac{\exp \!\left ( \!sim\!\left (f^{s}_{k},f^{t}_{k}  \right )\!/\tau  \!\right )}{\sum_{n=1}^{K}\! \exp\! \left (\! sim\!\left (f^{s}_{k},f^{t}_{n}  \right )\!/ \tau \!\right )}.\!
\end{gather}
$sim\!\left( \cdot \right)$ calculates the cosine similarity between segment and word representations. $\tau$ is temperature and we set $\tau\!=\!0.1$ following \citet{Wang_2021_CVPR}.

Overall, the total loss of expectation training is:
\begin{gather}
\mathcal{L}_{DiSeg}=\mathcal{L}_{mtl}+\mathcal{L}_{num}+\mathcal{L}_{ctr}.
\end{gather}

\subsection{Inference Policy}

\begin{algorithm}[t]
\small
\DontPrintSemicolon
  \KwInput{streaming speech inputs $\mathbf{s}$, lagging segments $k$ }
  \KwOutput{target outputs $\mathbf{\hat{y}}$}
  \KwInit{$\hat{y}_{0}\!=\!\left<\mathrm{BOS} \right>$, target index $t\!=\!1$, current received speech $\mathbf{\hat{s}}\!=\![\,]$}
  \BlankLine
  \While{$\;\hat{y}_{t-1}\!\neq\!\left<\mathrm{EOS} \right>$}{
  
    Extract speech features $\left\{a_{l} \right\}_{l=1}^{i}$ from $\mathbf{\hat{s}}$;\\
    Predict segmentation $\left\{b_{l} \right\}_{l=1}^{i}$;\\

    \eIf(\tcp*[f]{Translate}){$\;\;\sum_{l=1}^{i} b_{l}\!\geq\! t\!+\!k\!-\!1$ $\;\mathrm{or}\;$ $\mathbf{\hat{s}}\!==\!\mathbf{s}$ \\$\!\!$  }{
      Translate $\hat{y}_{t}$ based on $\mathbf{\hat{s}}$; $\;\;\;$$t\leftarrow t+1$;
    }
    (\tcp*[f]{Wait})
    {
      $\mathbf{\hat{s}}\leftarrow \mathbf{\hat{s}} + \mathbf{s}.\text{read()}$; \\
    }
    
    }
  \Return $\mathbf{\hat{y}}$;
\caption{Wait-seg Policy for DiSeg}
\label{algor}
\end{algorithm}

Owing to the proposed differentiable segmentation, the streaming speech inputs are divided into multiple segments, where each segment contains roughly one word. Accordingly, inspired by the wait-k policy \citep{ma-etal-2019-stacl} in simultaneous machine translation, we propose \emph{wait-seg policy} for DiSeg. Specifically, wait-seg policy first waits for $k$ segments, and then translates a target word whenever deciding to segment the streaming speech inputs, where $k$ is a hyperparameter to control the latency.
Formally, given lagging segments $k$, DiSeg translates $y_{t}$ when receiving $g\!\left ( t;k \right )$ speech features:
\begin{gather}
    g\!\left ( t;k \right )=\underset{i}{\mathrm{argmin}}\left ( \sum_{l=1}^{i} b_{l}\geq t+k-1 \right ).
\end{gather}
The specific inference is shown in Algorithm \ref{algor}.

To keep the training and inference matching, we also apply the wait-seg policy during training via the proposed wait-seg decoder. When translating $y_{t}$, wait-seg decoder will mask out the speech features $a_{i}$ that $i\!>\!g\!\left ( t;k \right )$ \citep{ma-etal-2019-stacl}. Accordingly, we introduce multi-task training and multi-latency training to enhance DiSeg performance.

\textbf{Multi-task Training} Since we adopt the multi-task learning framework (refer to Sec.\ref{sec:Background}), ASR and MT tasks should also adapt to DiSeg. Specifically, ASR task applies the same segmentation and policy (i.e., decoder) as the SimulST task, as both their inputs are speech. For the MT task, since the segment in the speech corresponds to the word in the transcription, MT task applies a uni-directional encoder and wait-k policy (i.e., decoder). Note that parameters of encoder and decoder are shared among various tasks. 

\textbf{Multi-latency Training} To enhance the DiSeg performance under multiple latency, we randomly sample $k$ from $\left [1,K  \right ]$ between batches during training \citep{multipath}. In inference, DiSeg only needs one model to complete SimulST under multiple arbitrary latency \citep{zhang-feng-2021-universal}, including offline speech translation (the latency is the complete speech duration). In this way, DiSeg develops a unified model that can handle both offline and simultaneous speech translation.

\section{Experiments}

\subsection{Datasets}

We conduct experiments on two end-to-end simultaneous translation benchmarks, MuST-C\footnote{\url{https://ict.fbk.eu/must-c}} English $\!\rightarrow\! $ German (En$\rightarrow$De, 234K pairs) and English $\!\rightarrow\! $ Spanish (En$\rightarrow$Es, 270K pairs) \cite{di-gangi-etal-2019-must}. We use \texttt{dev} as the validation set (1423 pairs for En$\rightarrow$De, 1316 pairs for En$\rightarrow$Es) and use \texttt{tst-COMMON} as the test set (2641 pairs for En$\rightarrow$De, 2502 pairs for En$\rightarrow$Es), respectively. 
For speech, we use the raw 16-bit 16kHz mono-channel audio wave. For text, we use SentencePiece \cite{kudo-richardson-2018-sentencepiece} to generate a unigram vocabulary of size $10000$, sharing between languages.
\begin{figure*}[t]
\centering
\subfigure[En$\rightarrow$De]{
\includegraphics[width=3.in]{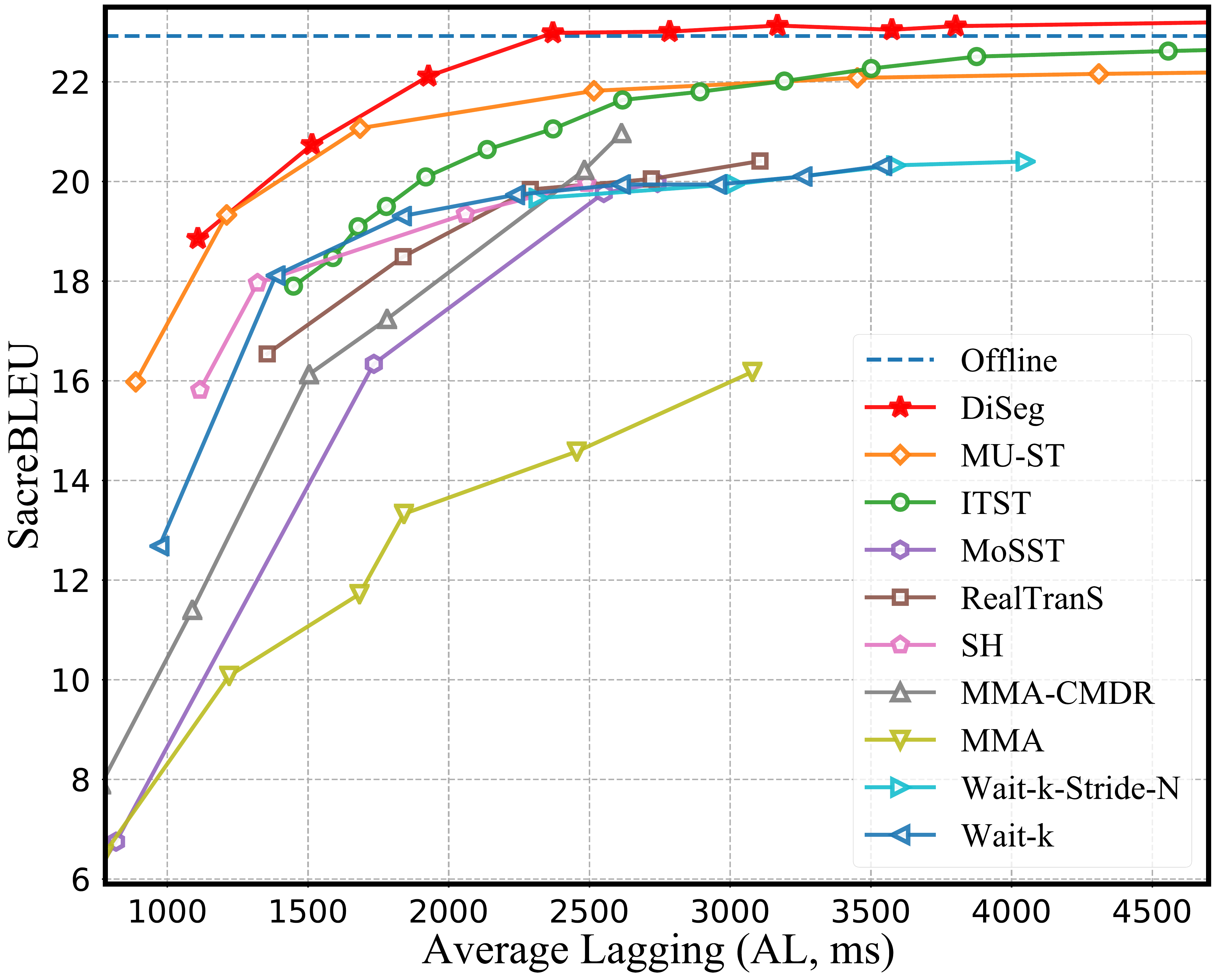}\label{fig:main1}
}
\subfigure[En$\rightarrow$Es]{
\includegraphics[width=3.in]{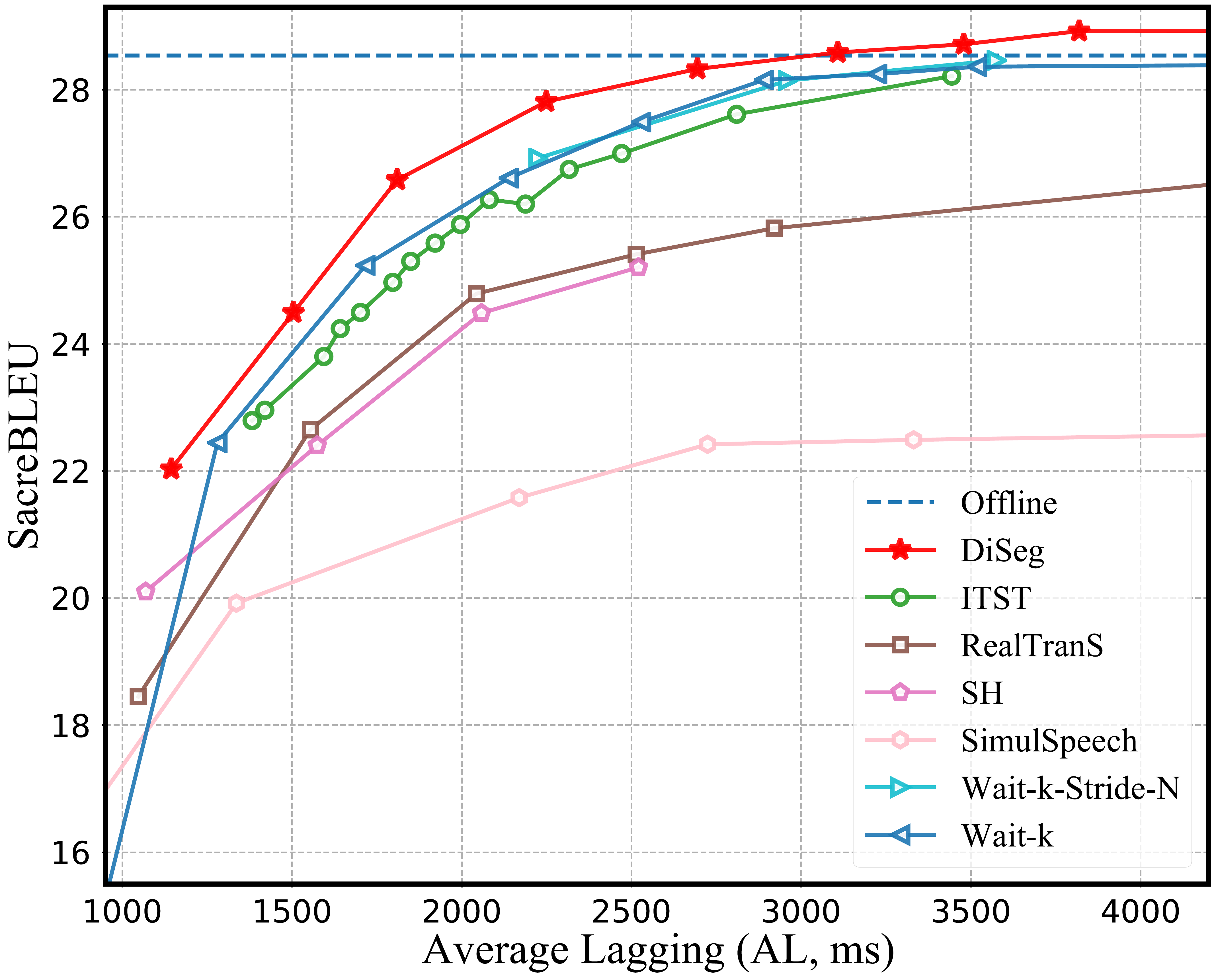}\label{fig:main2}
}

\caption{SimulST performance of translation quality against latency (AL, $ms$) on MuST-C En$\rightarrow$De and En$\rightarrow$Es.}
\label{fig:main}
\end{figure*}
\subsection{System Settings}
We conduct experiments on the following systems.

{\bf Offline} Offline speech translation, which waits for the complete speech inputs and then translates (bi-directional attention and greedy search).

{\bf Wait-k} Wait-k policy \citep{ma-etal-2019-stacl} with fixed segmentation of speech \citep{ma-etal-2020-simulmt}, which translates a word every 280$ms$.

{\bf Wait-k-Stride-n} A variation of wait-k policy \citep{zeng-etal-2021-realtrans}, which translates $n$ word every $n\!\times$280$ms$. We set $n\!=\!2$ following their best result.

{\bf MMA}\footnote{\url{https://github.com/pytorch/fairseq/tree/master/examples/simultaneous_translation}} Monotonic multihead attention \citep{Ma2019a}, which is adapted to SimulST by dividing speech into segments of equal length, i.e., 120$ms$, 200$ms$ and 280$ms\cdots$ \citep{ma-etal-2020-simulmt}.

{\bf MMA-CMDR} MMA with cross-modal decision regularization \citep{zaidi22_interspeech}, which leverages the transcription to improve the decision of MMA.

{\bf SimulSpeech} Segmentation based on word detector \citep{ren-etal-2020-simulspeech}, which also uses two knowledge distillations to improve the performance.

{\bf SH} Synchronized ASR-assisted SimulST \citep{chen-etal-2021-direct}, which uses the shortest hypothesis in ASR results to indicate the word number in speech.

{\bf RealTrans} A convolutional weighted-shrinking Transformer \citep{zeng-etal-2021-realtrans}, which detects the word number in the streaming speech and then decodes via the wait-k-stride-n policy.

{\bf MoSST}\footnote{\url{https://github.com/dqqcasia/mosst}} Monotonic-segmented streaming speech translation \cite{dong-etal-2022-learning}, which uses the integrate-and-firing method \citep{9054250} to segment the speech based on the cumulative acoustic information.

{\bf ITST}\footnote{\url{https://github.com/ictnlp/ITST}} Information-transport-based policy for SimulST \citep{ITST}, which quantifies the transported information from source to target, and then decides whether to translate according to the accumulated received information.

{\bf MU-ST} Segmentation based on the meaning unit \citep{zhang-etal-2022-learning}, which trains an external segmentation model based on the constructed data, and uses it to decide when to translate.

{\bf DiSeg} The proposed method in Sec.\ref{sec:method}.

All implementations are adapted from Fairseq Library \cite{ott-etal-2019-fairseq}. We use a pre-trained Wav2Vec2.0\footnote{\url{dl.fbaipublicfiles.com/fairseq/wav2vec/wav2vec_small.pt}} \citep{NEURIPS2020_92d1e1eb} as the acoustic feature extractor, and use a standard Transformer-Base \citep{NIPS2017_7181} as the translation model. For evaluation, we apply SimulEval\footnote{\url{https://github.com/facebookresearch/SimulEval}} \cite{ma-etal-2020-simuleval} to report SacreBLEU \citep{post-2018-call} for translation quality and Average Lagging (AL) \citep{ma-etal-2019-stacl} for latency. AL measures the average duration ($ms$) that target outputs lag behind the speech inputs. The calculation refer to Appendix \ref{app:numerical}.

\begin{figure*}[t]
\centering
\subfigure[Discreteness of $p_{i}$.]{
\includegraphics[width=1.45in]{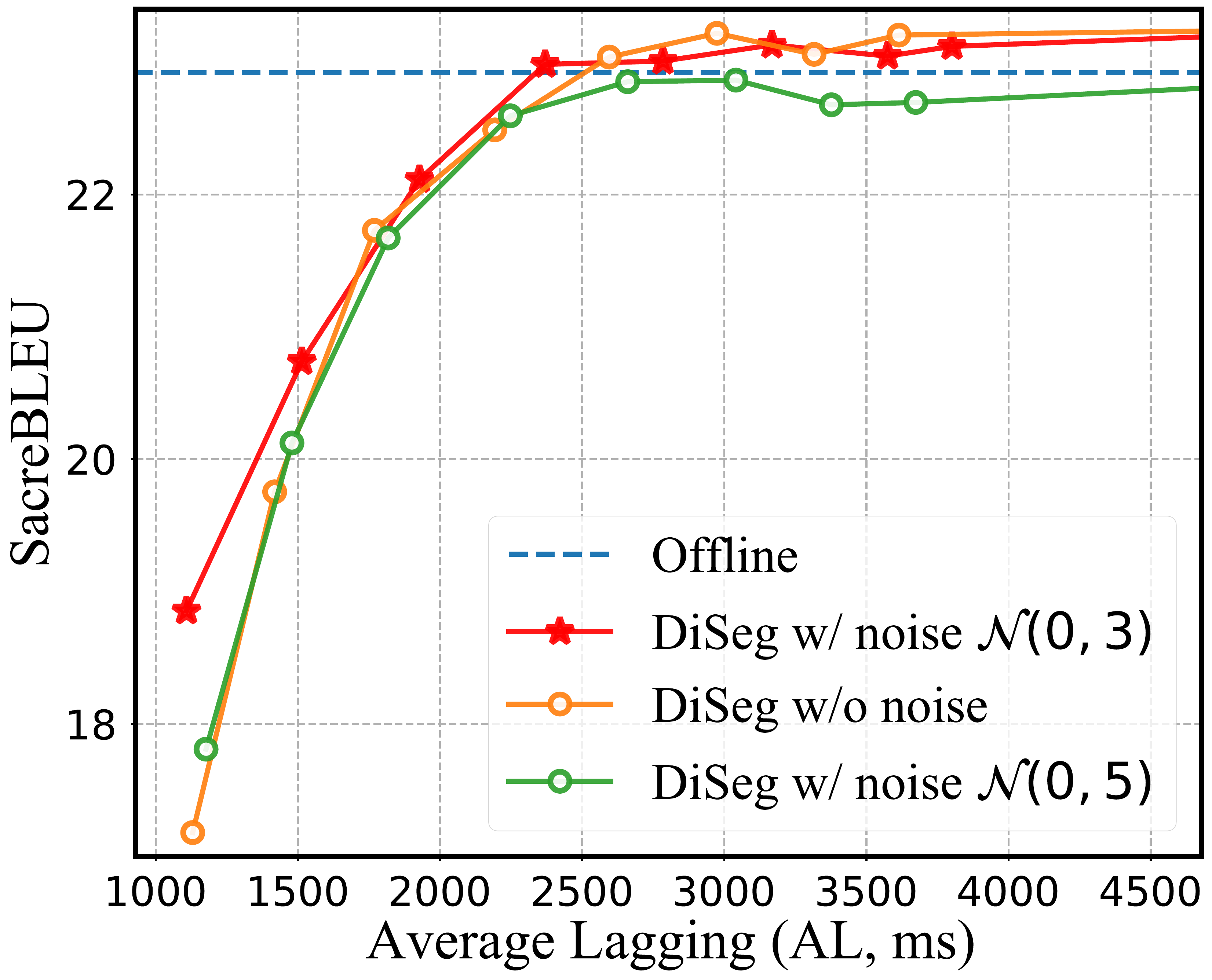}\label{fig:ablation1}
}
\subfigure[Number of segments.]{
\includegraphics[width=1.45in]{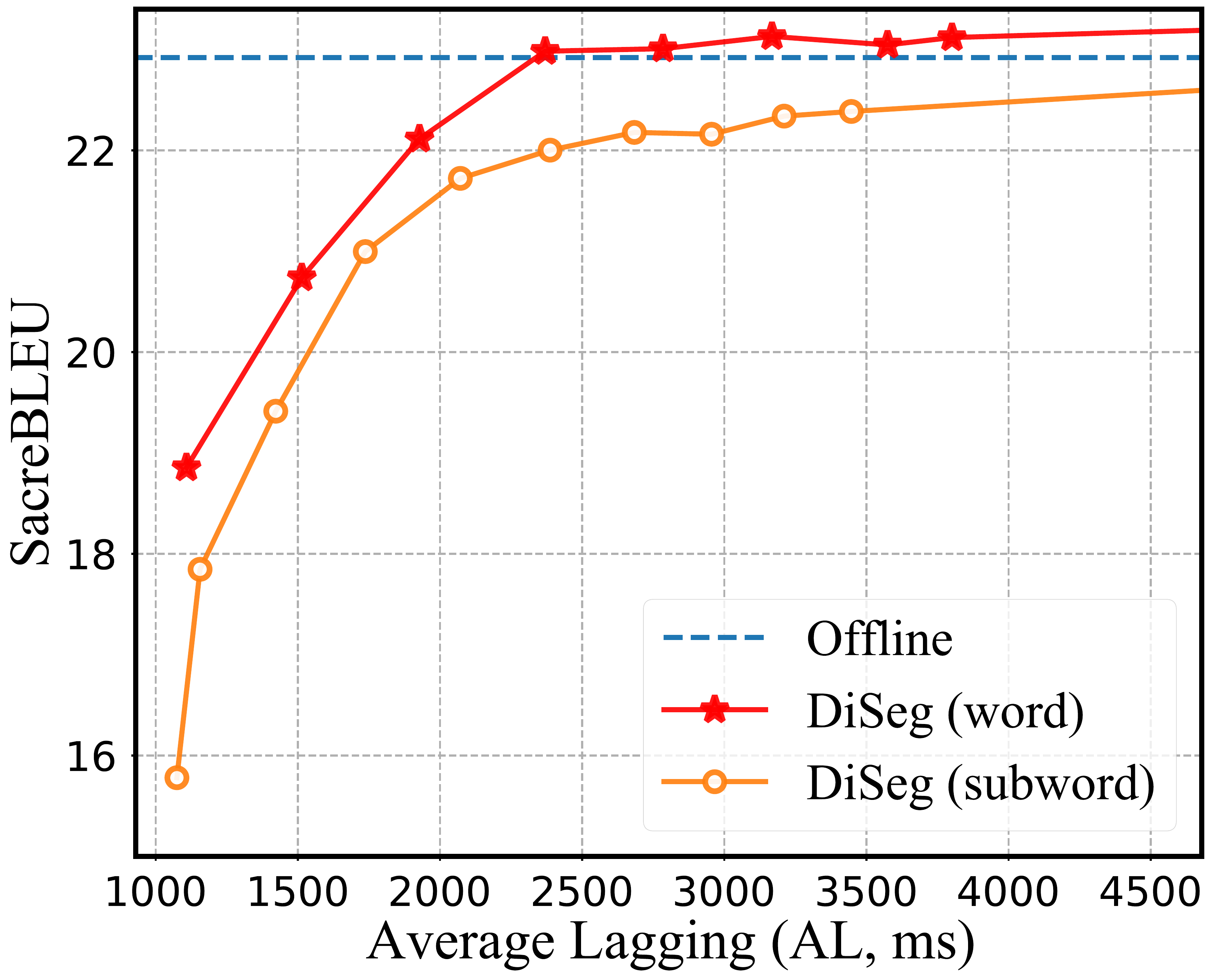}\label{fig:ablation2}
}
\subfigure[Learning of segmentation.]{
\includegraphics[width=1.45in]{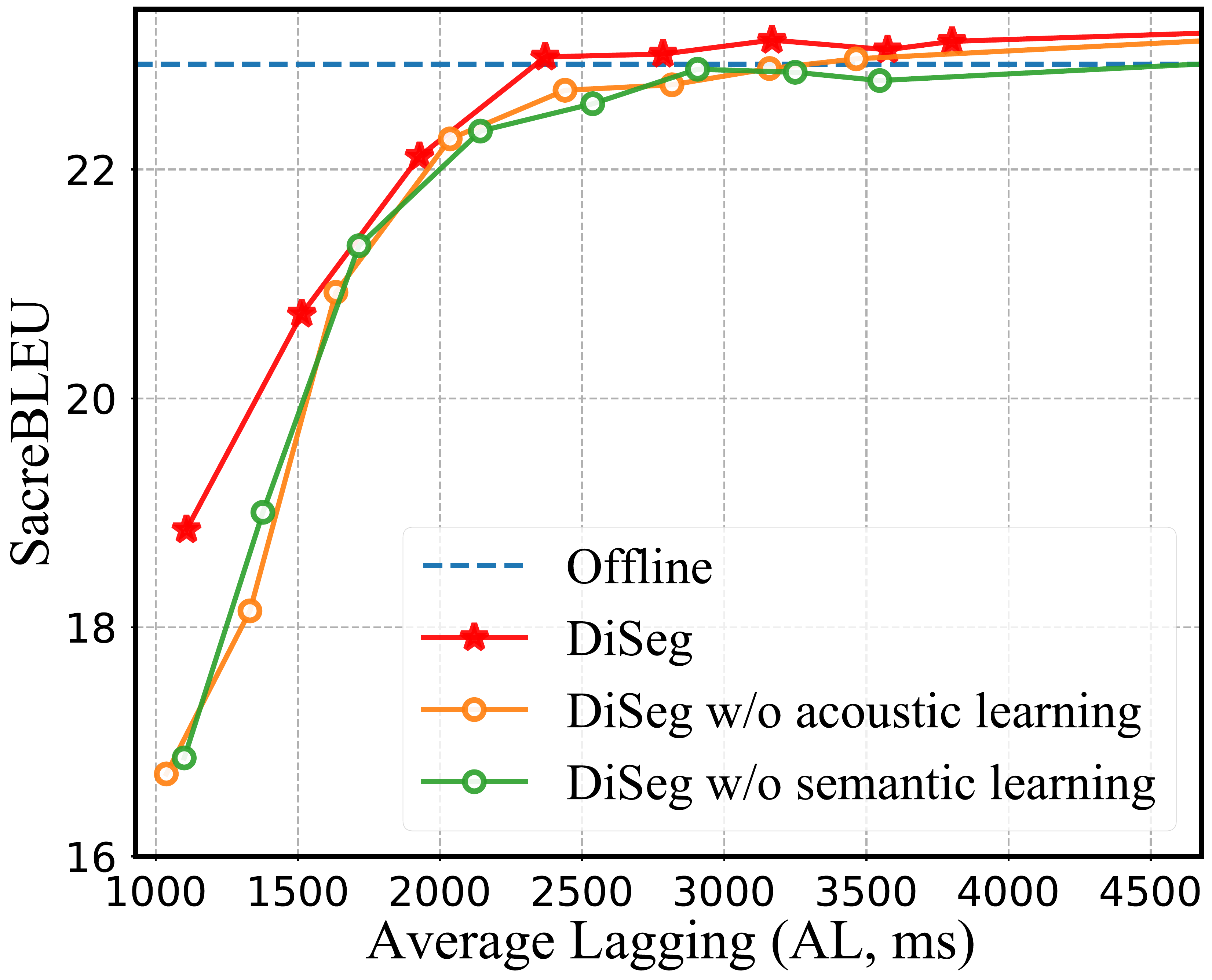}\label{fig:ablation3}
}
\subfigure[Wait-seg decoder.]{
\includegraphics[width=1.45in]{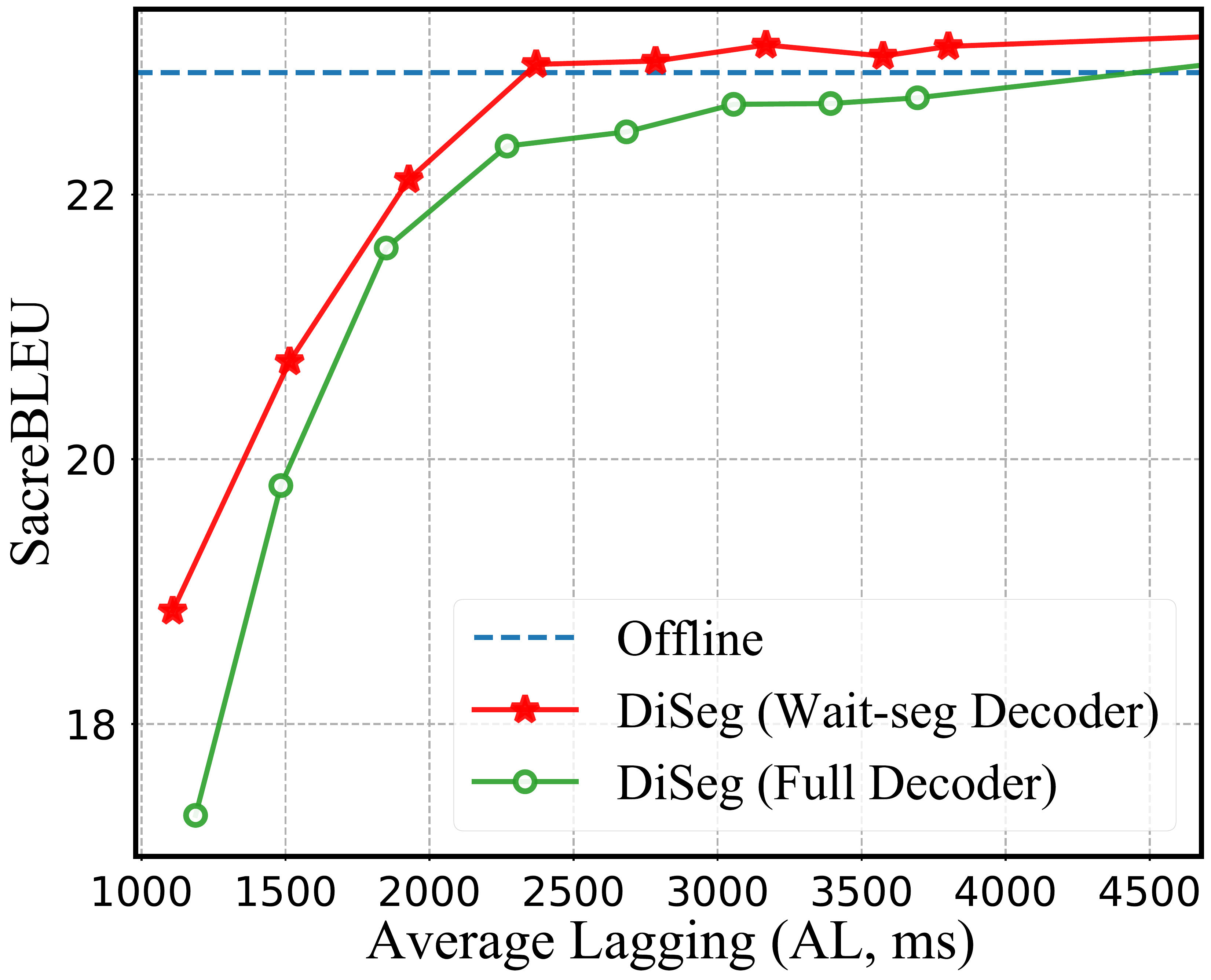}\label{fig:ablation4}
}
\caption{Ablation studies of DiSeg on MuST-C En$\rightarrow$De test set.}
\label{fig:ablation}
\end{figure*}

\subsection{Main Results}
We compare DiSeg and previous SimulST methods in Figure \ref{fig:main}, where we only train a single DiSeg model and adjusted the lagging number $k$ during the inference process to show the translation quality under different latency. Remarkably, DiSeg outperforms strong baselines under all latency and achieves state-of-the-art performance. Compared with fixed methods, such as Wait-k and MMA \citep{ma-etal-2020-simulmt}, DiSeg can dynamically decides segmentation according to the inputs instead of equal-length segmentation, which avoids breaking the acoustic integrity and thus achieves notable improvements. Compared with adaptive methods, including the state-of-the-art MU-ST, DiSeg also performs better. In previous adaptive methods, regardless of RealTrans, SH and SimulSpeech detecting the word number \citep{zeng-etal-2021-realtrans, chen-etal-2021-direct,ren-etal-2020-simulspeech}, MoSST and ITST comparing the acoustic information with a threshold \citep{dong-etal-2022-learning,ITST}, or MU-ST training an external segmentation model \citep{zhang-etal-2022-learning}, the final translation results is always non-differentiable to the segmentation, which hinders learning segmentation directly from the translation model. The proposed DiSeg turns the segmentation into differentiable, hence can learn translation-beneficial  segmentation directly from the translation model, thereby achieving better performance. Furthermore, unlike the previous methods using uni-directional (e.g., RealTrans and ITST) or bi-directional attention (e.g., MU-ST and MoSST), the proposed segmented attention can not only encode streaming inputs but also get comprehensive segment representations \citep{future-guided}. In particular, DiSeg achieves comparable performance with the offline model when lagging 2300$ms$ on En$\rightarrow$De and 3000$ms$ on En$\rightarrow$Es, which is attributed to the improvements on translation quality brought by the segmented attention.

\section{Analysis}

We conduct extensive analyses to study the effectiveness and specific improvements of DiSeg. Unless otherwise specified, all the results are reported on MuST-C En$\rightarrow$De test set.

\subsection{Ablation Study}

\textbf{Discreteness of Segmentation Probability} To make expectation training more suitable for inference, we encourage the discreteness of segment probability via introducing Gaussian noise $\mathcal{N}\!\left(0,n \right)$ in Eq.(\ref{eq:noise}). We compare the effect of discreteness in Figure \ref{fig:ablation1}, where appropriately encouraging discreteness effectively enhances expectation training, thereby improving DiSeg performance under low latency. However, too much noise will affect translation quality, especially under high latency, which is consistent with \citet{Arivazhagan2019}.

\textbf{Number of Segments} In DiSeg, we constrain the number of segments to be the word number $K$ in the transcription rather than subword. To verify the effectiveness of segmentation granularity, we compare different segment numbers in Figure \ref{fig:ablation2}, noting that $\mathcal{L}_{ctr}$ is also changed to be computed by subword embedding accordingly. Segmentation on word granularity is significantly better than subword granularity, mainly because many subwords are actually continuous and related in the speech, and segmentation at the word granularity can better preserve the acoustic integrity \citep{dong-etal-2022-learning}.

\textbf{Learning at Acoustic and Semantic Levels} DiSeg learns segmentation at the acoustic and semantic levels, so we show the effectiveness of acoustic and semantic learning in Figure \ref{fig:ablation3}. The results demonstrate that both acoustic and semantic learning play the important role in SimulST performance. Specifically, acoustic learning encourages related speech features to be in the same segment through expected segment attention, where ensuring acoustic integrity is more important for SimulST under low latency, thereby achieving an improvement of 2 BLEU (AL$\approx\!1500ms$). Semantic learning supervises the segment representations through the word representations in the transcription, which helps the segment representations to be more conducive to the translation model \citep{ye-etal-2022-cross}, thereby improving the translation quality.

\textbf{Effect of Wait-seg Decoder} DiSeg introduces wait-seg decoder to learn wait-seg policy. As shown in Figure \ref{fig:ablation4}, compared with full decoder that can focus on all speech features, wait-seg decoder enhances DiSeg's ability to translate based on partial speech \citep{ma-etal-2019-stacl} and thus achieves significant improvements during inference.

\subsection{Segmented Attention on Offline ST}

\begin{table}[t]
\centering
\small
\begin{tabular}{L{2.4cm}C{0.8cm}C{0.8cm}C{0.8cm}C{0.8cm}}\toprule
\multirow{2}{*}{\begin{tabular}[c]{@{}l@{}}\textbf{Encoder}\\ \textbf{Type}\end{tabular}}      & \multicolumn{2}{c}{\textbf{En}$\rightarrow$\textbf{De}} & \multicolumn{2}{c}{\textbf{En}$\rightarrow$\textbf{Es}} \\\cmidrule(lr){2-3}\cmidrule(lr){4-5}
      & Greedy       & Beam5      & Greedy       & Beam5      \\\midrule
Uni-directional  & 22.94        & 24.26      & 28.54        & 28.92      \\
Bi-directional   & 22.92        & 24.64      & 28.47        & 29.51      \\ 
DiSeg (segmented) & \textbf{23.34}        & \textbf{24.68}      & \textbf{28.96}        & \textbf{29.65}      \\
$\;\;-\mathcal{L}_{ctr}$  & 23.10        & 24.46      & 28.64        & 29.48     \\\bottomrule
\end{tabular}
\caption{Performance of segmented attention against bi-/uni-directional attention on offline speech translation.}
\label{tab:attn}
\end{table}

How to encode streaming inputs is an important concern for SimulST \citep{future-guided}, where offline translation uses bi-directional attention to encode the complete source input and existing SimulST methods always apply uni-directional attention \citep{zeng-etal-2021-realtrans,ITST}. DiSeg applies segmented attention, which consists of bi-directional attention within a segment and uni-directional attention between segments. To study the modeling capability of segmented attention, we compare the performance of uni-directional attention, bi-directional attention and DiSeg (segmented attention) on offline speech translation in Table \ref{tab:attn}. DiSeg is a unified model that can handle both simultaneous speech translation and offline speech translation together. Therefore, we employ the same model as SimulST to complete offline speech translation, while only setting $k=\infty$ and applying beam search during inference.

Uni- and bi-directional attention achieve similar performance in greedy search, which is consistent with \citet{pmlr-v139-wu21e}, while bi-directional attention performs better in beam search due to more comprehensive encoding. Owing to learning the translation-beneficial segmentation, DiSeg can outperform uni-/bi-directional attention on both greedy and beam search when only applying bi-directional attention within segments. Furthermore, when removing $\mathcal{L}_{ctr}$, segmented attention also achieves comparable performance to bi-directional attention, providing a new attention mechanism for future streaming models. Appendix \ref{sec:Visualization of Segmented Attention} gives visualization and more analyses of segmented attention.

\subsection{Segmentation Quality}
\label{sec:Segmentation Quality}

\begin{table}[t]
\centering
\small
\begin{tabular}{L{1.65cm}C{0.58cm}C{0.58cm}C{0.58cm}C{0.72cm}C{1.11cm}}\toprule
\textbf{Methods} & \textbf{P}($\uparrow$) & \textbf{R}($\uparrow$) & \textbf{F1}($\uparrow$)   & \textbf{OS}($0$) & \textbf{R-val}($\uparrow$) \\ \midrule
ES K-Means       & 30.7       & 18.0       & 22.7          & -41.2       & 39.7           \\
BES GMM          & 31.7       & 13.8       & 19.2          & -56.6       & 37.9           \\
VQ-CPC           & 18.2       & 54.1       & 27.3          & $\!$196.4       & $\!\!$-86.5          \\
VQ-VAE           & 16.4       & 56.8       & 25.5          & $\!$245.2       & $\!\!\!\!\!$-126.5         \\
SCPC             & 35.0       & 29.6       & 32.1          & -15.4       & 44.5           \\
DSegKNN          & 30.9       & 32.0       & 31.5          & $\;\;\;$3.5         & 40.7           \\\midrule
Fixed(280$ms$)    & 28.1       & 16.3       & 20.7          & -42.0       & 38.4           \\
DiSeg            & 34.9       & 32.3       & \textbf{33.5} & $\;\:$-7.4        & \textbf{44.6}  \\
$\;\;-\mathcal{L}_{ctr}$             & 33.9       & 31.0       & 32.4          & $\;\:$-8.5        & 43.9       \\\bottomrule   
\end{tabular}
\caption{Segmentation quality on Buckeye corpus. Better segmentation has OS close to 0, higher F1 and R-val.}
\label{tab:seg}
\end{table}

To explore the segmentation quality of DiSeg, we conduct experiments on speech segmentation task \citep{10.1007/3-540-46154-X_38} with the annotated Buckeye corpus\footnote{\url{https://buckeyecorpus.osu.edu}} \citep{PITT200589}. Table \ref{tab:seg} shows the segmentation performance of DiSeg and strong baselines \citep{8269008,KAMPER2017154,2020arXiv201207551K,9789954,fuchs2022unsupervised}, and the metrics include precision (P), recall (R), F1 score, over-segmentation (OS) and R-value (refer to Appendix \ref{sec:Metrics of Word Segmentation Task} for the calculation). The results show that DiSeg achieves better segmentation performance and $\mathcal{L}_{ctr}$ can improve the segmentation quality by 1\% score. More importantly, DiSeg achieves an OS score close to $0$, demonstrating that DiSeg can get the appropriate number of segments, thereby avoiding too many or too few segments affecting SimulST performance \citep{wait-info}. We will analyze the number of segments following.

\subsection{Segmentation Quantity}

\begin{figure}[t]
\centering
    \includegraphics[width=2.6in]{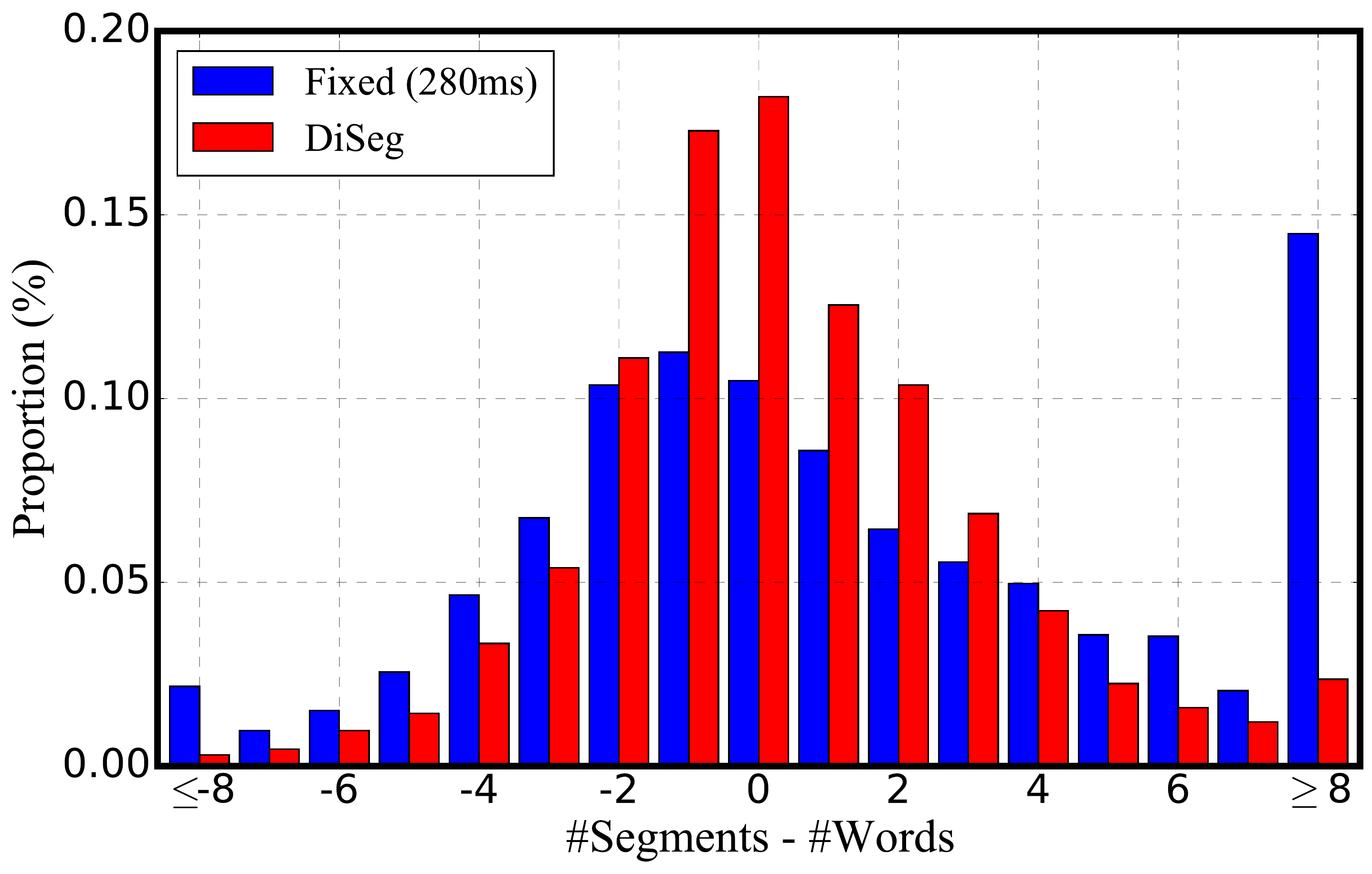}
    \caption{Distribution of the difference between the segment number generated by DiSeg and the word number in transcription, evaluated on MuST-C En$\rightarrow$De.}
    \label{fig:num}
\end{figure}

In training, we constrain the number of segments to be the word number in the transcription. To verify its effectiveness, we count the difference between the segment number and the word number during inference (i.e., \#Segments$-$\#Words) in Figure \ref{fig:num}. Compared with the previous work considering that 280$ms$ corresponds to a word on average \citep{ma-etal-2020-simulmt,zaidi22_interspeech}, DiSeg can get a more accurate number of segments, where the difference between the segment number and the word number is less than $2$ in 70\% of cases. Besides, as reported in Table \ref{tab:seg}, the OS score on the automatic speech segmentation task also demonstrates that DiSeg can achieve an appropriate number of segments. Therefore, constraining the expected segment number in expectation training is effective to control the number of segments.

\subsection{Adapting Multi-task Learning to SimulST}
During training, we adjust ASR task (segmented encoder + wait-seg decoder) and MT task (uni-encoder + wait-k decoder) in multi-task learning to adapt to DiSeg, so we verify the effectiveness of the adaptation in Table \ref{tab:mtl}. In the proposed adaptation \#1, since a speech segment corresponds to a word, both uni-encoder and wait-k decoder in MT task are fully compatible with the segmented attention and wait-seg decoder in SimulST task, thereby performing best. Both using bi-encoder (i.e., setting \#5-7) or full decoder (i.e., setting \#2-4) in ASR and MT tasks will affect DiSeg performance, and the performance degradation caused by the encoder mismatch is more serious. In general, ASR and MT tasks should be consistent and compatible with the SimulST task when adapting multi-task learning.

\begin{table}[t]
\small
\begin{tabular}{ccccccc} \toprule
\multirow{2}{*}{\textbf{\#}} & \multicolumn{2}{c}{\textbf{ASR}} & \multicolumn{2}{c}{\textbf{MT}} & \multirow{2}{*}{\textbf{AL}} & \multirow{2}{*}{\textbf{BLEU}} \\ \cmidrule(lr){2-3}\cmidrule(lr){4-5}
                    & Enc.     & Dec.         & Enc.      & Dec.       &                     &                       \\ \midrule
1                   & \textbf{seg}     & \textbf{wait-seg}     & \textbf{uni}       & \textbf{wait-k}     & \textbf{1514}                & \textbf{20.74}                \\
2                   & seg     & wait-seg     & uni       & full       & 1428                & 19.32                 \\
3                   & seg     & full         & uni       & wait-k     & 1404                & 19.58                 \\
4                   & seg     & full         & uni       & full       & 1398                & 19.39                 \\\midrule
5                   & seg     & wait-seg     & bi        & full       & 1416                & 18.93                 \\
6                   & bi       & full         & uni       & wait-k     & 1704                & 20.23                 \\
7                   & bi       & full         & bi        & full       & 1374                & 18.82                \\\bottomrule   
\end{tabular}
\caption{Performance with various settings of multi-task learning, evaluating with $k\!=\!3$ on MuST-C En$\rightarrow$De.}
\label{tab:mtl}
\end{table}

\section{Related Work}

Early SimulST methods segment speech and then use the cascaded model (ASR+MT) to translate each segment \citep{10.2307/30219116,yarmohammadi-etal-2013-incremental,rangarajan-sridhar-etal-2013-segmentation,zhang2023hidden,bs}.
Recent end-to-end SimulST methods fall into fixed and adaptive. For fixed, \citet{ma-etal-2020-simulmt} proposed fixed pre-decision to divide speech into equal-length segments, and migrated simultaneous MT method, such as wait-k \citep{ma-etal-2019-stacl,zhang-feng-2021-universal,zhang-feng-2021-icts,post-eval} and MMA \citep{Ma2019a}, to SimulST. For adaptive, \citet{ren-etal-2020-simulspeech} proposed SimulSpeech to detect the word in speech. \citet{chen-etal-2021-direct} used ASR results to indicate the word number. \citet{zeng-etal-2021-realtrans} proposed RealTrans, which detects the source word and further shrinks the speech length. \citet{dong-etal-2022-learning} proposed MoSST to translate after the acoustic information exceeding 1. \citet{ITST} proposed ITST to judge whether the received information is sufficient for translation. \citet{zhang-etal-2022-learning} proposed MU-ST, which constructs the segmentation labels based on meaning unit, and uses it to train a segmentation model.

In the previous method, whether using an external segmentation model or the detector, the segmentation cannot receive the gradient (i.e., learning) from the underlying translation model as the hard segmentation is not differentiable. Owing to the differentiable property, DiSeg can be jointly trained with the underlying translation model and directly learn translation-beneficial segmentation.

\section{Conclusion}
In this study, we propose differentiable segmentation (DiSeg) for simultaneous speech translation to directly learn segmentation from the underlying translation model. Experiments show the superiority of DiSeg in terms of SimulST performance, attention mechanism and segmentation quality.

Future researches will delve into the untapped potential of differentiable segmentation in such streaming models and long sequence modeling, thereby reducing feedback latency or computational cost without compromising performance.

\section*{Acknowledgements}
We thank all the anonymous reviewers for their insightful and valuable comments. This work was supported by National Key R\&D Program of China (NO. 2018AAA0102502).

\section*{Limitations}

In this study, we propose differentiable segmentation to learn how to segment speech from the underlying translation model, and verify its effectiveness on simultaneous speech translation. However, since it can be jointly trained with the underlying task (sequence-to-sequence task), differentiable segmentation is not limited to the SimulST task, but can be generalized to more streaming/online tasks, such as streaming automatic speech recognition (streaming ASR), simultaneous machine translation (SiMT), real-time text-to-speech synthesis (real-time TTS), online tagging and streaming parsing. Given that there may be some task-specific differences between various tasks, this work only focuses on the differentiable segmentation in the SimulST task, and we leave the study of how to apply differentiable segmentation to other streaming tasks into our future work.

\bibliography{custom}

\begin{thebibliography}{78}
\expandafter\ifx\csname natexlab\endcsname\relax\def\natexlab#1{#1}\fi

\bibitem[{Anastasopoulos and Chiang(2018)}]{anastasopoulos-chiang-2018-tied}
Antonios Anastasopoulos and David Chiang. 2018.
\newblock \href {https://doi.org/10.18653/v1/N18-1008} {Tied multitask learning
  for neural speech translation}.
\newblock In \emph{Proceedings of the 2018 Conference of the North {A}merican
  Chapter of the Association for Computational Linguistics: Human Language
  Technologies, Volume 1 (Long Papers)}, pages 82--91, New Orleans, Louisiana.
  Association for Computational Linguistics.

\bibitem[{Arivazhagan et~al.(2019)Arivazhagan, Cherry, Macherey, Chiu, Yavuz,
  Pang, Li, and Raffel}]{Arivazhagan2019}
Naveen Arivazhagan, Colin Cherry, Wolfgang Macherey, Chung-cheng Chiu, Semih
  Yavuz, Ruoming Pang, Wei Li, and Colin Raffel. 2019.
\newblock \href {https://doi.org/10.18653/v1/p19-1126} {{Monotonic Infinite
  Lookback Attention for Simultaneous Machine Translation}}.
\newblock pages 1313--1323.

\bibitem[{Baevski et~al.(2020)Baevski, Zhou, Mohamed, and
  Auli}]{NEURIPS2020_92d1e1eb}
Alexei Baevski, Yuhao Zhou, Abdelrahman Mohamed, and Michael Auli. 2020.
\newblock \href
  {https://proceedings.neurips.cc/paper/2020/file/92d1e1eb1cd6f9fba3227870bb6d7f07-Paper.pdf}
  {wav2vec 2.0: A framework for self-supervised learning of speech
  representations}.
\newblock In \emph{Advances in Neural Information Processing Systems},
  volume~33, pages 12449--12460. Curran Associates, Inc.

\bibitem[{Bhati et~al.(2022)Bhati, Villalba, Żelasko, Moro-Velazquez, and
  Dehak}]{9789954}
Saurabhchand Bhati, Jesús Villalba, Piotr Żelasko, Laureano Moro-Velazquez,
  and Najim Dehak. 2022.
\newblock \href {https://doi.org/10.1109/TASLP.2022.3180684} {Unsupervised
  speech segmentation and variable rate representation learning using segmental
  contrastive predictive coding}.
\newblock \emph{IEEE/ACM Transactions on Audio, Speech, and Language
  Processing}, 30:2002--2014.

\bibitem[{Chen et~al.(2021)Chen, Ma, Zheng, and Huang}]{chen-etal-2021-direct}
Junkun Chen, Mingbo Ma, Renjie Zheng, and Liang Huang. 2021.
\newblock \href {https://doi.org/10.18653/v1/2021.findings-acl.406} {Direct
  simultaneous speech-to-text translation assisted by synchronized streaming
  {ASR}}.
\newblock In \emph{Findings of the Association for Computational Linguistics:
  ACL-IJCNLP 2021}, pages 4618--4624, Online. Association for Computational
  Linguistics.

\bibitem[{Cho and Esipova(2016)}]{Cho2016}
Kyunghyun Cho and Masha Esipova. 2016.
\newblock \href {http://arxiv.org/abs/1606.02012} {{Can neural machine
  translation do simultaneous translation?}}

\bibitem[{Demuynck and Laureys(2002)}]{10.1007/3-540-46154-X_38}
Kris Demuynck and Tom Laureys. 2002.
\newblock \href {https://link.springer.com/chapter/10.1007/3-540-46154-X_38} {A
  comparison of different approaches to automatic speech segmentation}.
\newblock In \emph{Text, Speech and Dialogue}, pages 277--284, Berlin,
  Heidelberg. Springer Berlin Heidelberg.

\bibitem[{Di~Gangi et~al.(2019)Di~Gangi, Cattoni, Bentivogli, Negri, and
  Turchi}]{di-gangi-etal-2019-must}
Mattia~A. Di~Gangi, Roldano Cattoni, Luisa Bentivogli, Matteo Negri, and Marco
  Turchi. 2019.
\newblock \href {https://doi.org/10.18653/v1/N19-1202} {{M}u{ST}-{C}: a
  {M}ultilingual {S}peech {T}ranslation {C}orpus}.
\newblock In \emph{Proceedings of the 2019 Conference of the North {A}merican
  Chapter of the Association for Computational Linguistics: Human Language
  Technologies, Volume 1 (Long and Short Papers)}, pages 2012--2017,
  Minneapolis, Minnesota. Association for Computational Linguistics.

\bibitem[{Ding et~al.(2019)Ding, Xu, and Koehn}]{ding-etal-2019-saliency}
Shuoyang Ding, Hainan Xu, and Philipp Koehn. 2019.
\newblock \href {https://doi.org/10.18653/v1/W19-5201} {Saliency-driven word
  alignment interpretation for neural machine translation}.
\newblock In \emph{Proceedings of the Fourth Conference on Machine Translation
  (Volume 1: Research Papers)}, pages 1--12, Florence, Italy. Association for
  Computational Linguistics.

\bibitem[{Dong and Xu(2020)}]{9054250}
Linhao Dong and Bo~Xu. 2020.
\newblock \href {https://doi.org/10.1109/ICASSP40776.2020.9054250} {Cif:
  Continuous integrate-and-fire for end-to-end speech recognition}.
\newblock In \emph{ICASSP 2020 - 2020 IEEE International Conference on
  Acoustics, Speech and Signal Processing (ICASSP)}, pages 6079--6083.

\bibitem[{Dong et~al.(2022)Dong, Zhu, Wang, and Li}]{dong-etal-2022-learning}
Qian Dong, Yaoming Zhu, Mingxuan Wang, and Lei Li. 2022.
\newblock \href {https://doi.org/10.18653/v1/2022.acl-long.50} {Learning when
  to translate for streaming speech}.
\newblock In \emph{Proceedings of the 60th Annual Meeting of the Association
  for Computational Linguistics (Volume 1: Long Papers)}, pages 680--694,
  Dublin, Ireland. Association for Computational Linguistics.

\bibitem[{Elbayad et~al.(2020)Elbayad, Besacier, and Verbeek}]{multipath}
Maha Elbayad, Laurent Besacier, and Jakob Verbeek. 2020.
\newblock \href {https://doi.org/10.21437/Interspeech.2020-1241} {{Efficient
  Wait-k Models for Simultaneous Machine Translation}}.
\newblock In \emph{Proc. Interspeech 2020}, pages 1461--1465.

\bibitem[{Foerster et~al.(2016)Foerster, Assael, de~Freitas, and
  Whiteson}]{NIPS2016_c7635bfd}
Jakob Foerster, Ioannis~Alexandros Assael, Nando de~Freitas, and Shimon
  Whiteson. 2016.
\newblock \href
  {https://proceedings.neurips.cc/paper/2016/file/c7635bfd99248a2cdef8249ef7bfbef4-Paper.pdf}
  {Learning to communicate with deep multi-agent reinforcement learning}.
\newblock In \emph{Advances in Neural Information Processing Systems},
  volume~29. Curran Associates, Inc.

\bibitem[{Fuchs et~al.(2022)Fuchs, Hoshen, and Keshet}]{fuchs2022unsupervised}
Tzeviya~Sylvia Fuchs, Yedid Hoshen, and Joseph Keshet. 2022.
\newblock \href {https://arxiv.org/abs/2204.13094} {Unsupervised word
  segmentation using k nearest neighbors}.
\newblock \emph{arXiv preprint arXiv:2204.13094}.

\bibitem[{Fügen et~al.(2007)Fügen, Waibel, and Kolss}]{10.2307/30219116}
Christian Fügen, Alex Waibel, and Muntsin Kolss. 2007.
\newblock \href {https://link.springer.com/article/10.1007/s10590-008-9047-0}
  {Simultaneous translation of lectures and speeches}.
\newblock \emph{Machine Translation}, 21(4):209--252.

\bibitem[{Gu et~al.(2017)Gu, Neubig, Cho, and Li}]{gu-etal-2017-learning}
Jiatao Gu, Graham Neubig, Kyunghyun Cho, and Victor~O.K. Li. 2017.
\newblock \href {https://www.aclweb.org/anthology/E17-1099} {Learning to
  translate in real-time with neural machine translation}.
\newblock In \emph{Proceedings of the 15th Conference of the {E}uropean Chapter
  of the Association for Computational Linguistics: Volume 1, Long Papers},
  pages 1053--1062, Valencia, Spain. Association for Computational Linguistics.

\bibitem[{Guo et~al.(2022)Guo, Zhang, and Feng}]{post-eval}
Shoutao Guo, Shaolei Zhang, and Yang Feng. 2022.
\newblock \href {https://aclanthology.org/2022.findings-emnlp.167} {Turning
  fixed to adaptive: Integrating post-evaluation into simultaneous machine
  translation}.
\newblock In \emph{Findings of the Association for Computational Linguistics:
  EMNLP 2022}, pages 2264--2278, Abu Dhabi, United Arab Emirates. Association
  for Computational Linguistics.

\bibitem[{Guo et~al.(2023)Guo, Zhang, and Feng}]{bs}
Shoutao Guo, Shaolei Zhang, and Yang Feng. 2023.
\newblock \href {https://arxiv.org/abs/2305.12774} {Learning optimal policy for
  simultaneous machine translation via binary search}.
\newblock In \emph{Proceedings of the 61th Annual Meeting of the Association
  for Computational Linguistics}. Association for Computational Linguistics.

\bibitem[{Hamon et~al.(2009)Hamon, F{\"u}gen, Mostefa, Arranz, Kolss, Waibel,
  and Choukri}]{hamon-etal-2009-end}
Olivier Hamon, Christian F{\"u}gen, Djamel Mostefa, Victoria Arranz, Muntsin
  Kolss, Alex Waibel, and Khalid Choukri. 2009.
\newblock \href {https://aclanthology.org/E09-1040} {End-to-end evaluation in
  simultaneous translation}.
\newblock In \emph{Proceedings of the 12th Conference of the {E}uropean Chapter
  of the {ACL} ({EACL} 2009)}, pages 345--353, Athens, Greece. Association for
  Computational Linguistics.

\bibitem[{Iranzo~Sanchez et~al.(2022)Iranzo~Sanchez, Civera, and
  Juan-C{\'\i}scar}]{iranzo-sanchez-etal-2022-simultaneous}
Javier Iranzo~Sanchez, Jorge Civera, and Alfons Juan-C{\'\i}scar. 2022.
\newblock \href {https://doi.org/10.18653/v1/2022.acl-long.480} {From
  simultaneous to streaming machine translation by leveraging streaming
  history}.
\newblock In \emph{Proceedings of the 60th Annual Meeting of the Association
  for Computational Linguistics (Volume 1: Long Papers)}, pages 6972--6985,
  Dublin, Ireland. Association for Computational Linguistics.

\bibitem[{Kamper et~al.(2017{\natexlab{a}})Kamper, Jansen, and
  Goldwater}]{KAMPER2017154}
Herman Kamper, Aren Jansen, and Sharon Goldwater. 2017{\natexlab{a}}.
\newblock \href {https://doi.org/https://doi.org/10.1016/j.csl.2017.04.008} {A
  segmental framework for fully-unsupervised large-vocabulary speech
  recognition}.
\newblock \emph{Computer Speech \& Language}, 46:154--174.

\bibitem[{Kamper et~al.(2017{\natexlab{b}})Kamper, Livescu, and
  Goldwater}]{8269008}
Herman Kamper, Karen Livescu, and Sharon Goldwater. 2017{\natexlab{b}}.
\newblock \href {https://doi.org/10.1109/ASRU.2017.8269008} {An embedded
  segmental k-means model for unsupervised segmentation and clustering of
  speech}.
\newblock In \emph{2017 IEEE Automatic Speech Recognition and Understanding
  Workshop (ASRU)}, pages 719--726.

\bibitem[{{Kamper} and {van Niekerk}(2020)}]{2020arXiv201207551K}
Herman {Kamper} and Benjamin {van Niekerk}. 2020.
\newblock \href {http://arxiv.org/abs/2012.07551} {{Towards unsupervised phone
  and word segmentation using self-supervised vector-quantized neural
  networks}}.
\newblock \emph{arXiv e-prints}, page arXiv:2012.07551.

\bibitem[{Kudo and Richardson(2018)}]{kudo-richardson-2018-sentencepiece}
Taku Kudo and John Richardson. 2018.
\newblock \href {https://doi.org/10.18653/v1/D18-2012} {{S}entence{P}iece: A
  simple and language independent subword tokenizer and detokenizer for neural
  text processing}.
\newblock In \emph{Proceedings of the 2018 Conference on Empirical Methods in
  Natural Language Processing: System Demonstrations}, pages 66--71, Brussels,
  Belgium. Association for Computational Linguistics.

\bibitem[{K{\"u}rzinger et~al.(2020)K{\"u}rzinger, Winkelbauer, Li, Watzel, and
  Rigoll}]{10.1007/978-3-030-60276-5_27}
Ludwig K{\"u}rzinger, Dominik Winkelbauer, Lujun Li, Tobias Watzel, and Gerhard
  Rigoll. 2020.
\newblock \href
  {https://link.springer.com/chapter/10.1007/978-3-030-60276-5_27}
  {Ctc-segmentation of large corpora for german end-to-end speech recognition}.
\newblock In \emph{Speech and Computer}, pages 267--278, Cham. Springer
  International Publishing.

\bibitem[{Le-Khac et~al.(2020)Le-Khac, Healy, and Smeaton}]{9226466}
Phuc~H. Le-Khac, Graham Healy, and Alan~F. Smeaton. 2020.
\newblock \href {https://doi.org/10.1109/ACCESS.2020.3031549} {Contrastive
  representation learning: A framework and review}.
\newblock \emph{IEEE Access}, 8:193907--193934.

\bibitem[{Liang et~al.(2021)Liang, Xu, and Zhang}]{liang2021transformer}
Chengdong Liang, Menglong Xu, and Xiao-Lei Zhang. 2021.
\newblock \href
  {https://ui.adsabs.harvard.edu/abs/2021arXiv210315722L/abstract}
  {Transformer-based end-to-end speech recognition with residual gaussian-based
  self-attention}.
\newblock \emph{arXiv preprint arXiv:2103.15722}.

\bibitem[{Liu et~al.(2020)Liu, Zhu, Zhang, and Zong}]{liu2020bridging}
Yuchen Liu, Junnan Zhu, Jiajun Zhang, and Chengqing Zong. 2020.
\newblock \href {https://arxiv.org/abs/2010.14920} {Bridging the modality gap
  for speech-to-text translation}.
\newblock \emph{arXiv preprint arXiv:2010.14920}.

\bibitem[{Luong et~al.(2015)Luong, Pham, and
  Manning}]{luong-etal-2015-effective}
Thang Luong, Hieu Pham, and Christopher~D. Manning. 2015.
\newblock \href {https://doi.org/10.18653/v1/D15-1166} {Effective approaches to
  attention-based neural machine translation}.
\newblock In \emph{Proceedings of the 2015 Conference on Empirical Methods in
  Natural Language Processing}, pages 1412--1421, Lisbon, Portugal. Association
  for Computational Linguistics.

\bibitem[{Ma et~al.(2019)Ma, Huang, Xiong, Zheng, Liu, Zheng, Zhang, He, Liu,
  Li, Wu, and Wang}]{ma-etal-2019-stacl}
Mingbo Ma, Liang Huang, Hao Xiong, Renjie Zheng, Kaibo Liu, Baigong Zheng,
  Chuanqiang Zhang, Zhongjun He, Hairong Liu, Xing Li, Hua Wu, and Haifeng
  Wang. 2019.
\newblock \href {https://doi.org/10.18653/v1/P19-1289} {{STACL}: Simultaneous
  translation with implicit anticipation and controllable latency using
  prefix-to-prefix framework}.
\newblock In \emph{Proceedings of the 57th Annual Meeting of the Association
  for Computational Linguistics}, pages 3025--3036, Florence, Italy.
  Association for Computational Linguistics.

\bibitem[{Ma et~al.(2020{\natexlab{a}})Ma, Dousti, Wang, Gu, and
  Pino}]{ma-etal-2020-simuleval}
Xutai Ma, Mohammad~Javad Dousti, Changhan Wang, Jiatao Gu, and Juan Pino.
  2020{\natexlab{a}}.
\newblock \href {https://doi.org/10.18653/v1/2020.emnlp-demos.19} {{SIMULEVAL}:
  An evaluation toolkit for simultaneous translation}.
\newblock In \emph{Proceedings of the 2020 Conference on Empirical Methods in
  Natural Language Processing: System Demonstrations}, pages 144--150, Online.
  Association for Computational Linguistics.

\bibitem[{Ma et~al.(2020{\natexlab{b}})Ma, Pino, and
  Koehn}]{ma-etal-2020-simulmt}
Xutai Ma, Juan Pino, and Philipp Koehn. 2020{\natexlab{b}}.
\newblock \href {https://aclanthology.org/2020.aacl-main.58} {{S}imul{MT} to
  {S}imul{ST}: Adapting simultaneous text translation to end-to-end
  simultaneous speech translation}.
\newblock In \emph{Proceedings of the 1st Conference of the Asia-Pacific
  Chapter of the Association for Computational Linguistics and the 10th
  International Joint Conference on Natural Language Processing}, pages
  582--587, Suzhou, China. Association for Computational Linguistics.

\bibitem[{Ma et~al.(2020{\natexlab{c}})Ma, Pino, Cross, Puzon, and
  Gu}]{Ma2019a}
Xutai Ma, Juan~Miguel Pino, James Cross, Liezl Puzon, and Jiatao Gu.
  2020{\natexlab{c}}.
\newblock \href {https://openreview.net/forum?id=Hyg96gBKPS} {Monotonic
  multihead attention}.
\newblock In \emph{International Conference on Learning Representations}.

\bibitem[{Ma et~al.(2021)Ma, Wang, Dousti, Koehn, and Pino}]{9414897}
Xutai Ma, Yongqiang Wang, Mohammad~Javad Dousti, Philipp Koehn, and Juan Pino.
  2021.
\newblock \href {https://doi.org/10.1109/ICASSP39728.2021.9414897} {Streaming
  simultaneous speech translation with augmented memory transformer}.
\newblock In \emph{ICASSP 2021 - 2021 IEEE International Conference on
  Acoustics, Speech and Signal Processing (ICASSP)}, pages 7523--7527.

\bibitem[{Nguyen et~al.(2020)Nguyen, Bougares, Tomashenko, Est{\`e}ve, and
  laurent besacier}]{nguyen2020investigating}
Ha~Nguyen, Fethi Bougares, Natalia Tomashenko, Yannick Est{\`e}ve, and laurent
  besacier. 2020.
\newblock \href {https://openreview.net/forum?id=SR2L__h9q9p} {Investigating
  self-supervised pre-training for end-to-end speech translation}.
\newblock In \emph{ICML 2020 Workshop on Self-supervision in Audio and Speech}.

\bibitem[{Nguyen et~al.(2021)Nguyen, Estève, and Besacier}]{9414276}
Ha~Nguyen, Yannick Estève, and Laurent Besacier. 2021.
\newblock \href {https://doi.org/10.1109/ICASSP39728.2021.9414276} {An
  empirical study of end-to-end simultaneous speech translation decoding
  strategies}.
\newblock In \emph{ICASSP 2021 - 2021 IEEE International Conference on
  Acoustics, Speech and Signal Processing (ICASSP)}, pages 7528--7532.

\bibitem[{Oda et~al.(2014)Oda, Neubig, Sakti, Toda, and
  Nakamura}]{oda-etal-2014-optimizing}
Yusuke Oda, Graham Neubig, Sakriani Sakti, Tomoki Toda, and Satoshi Nakamura.
  2014.
\newblock \href {https://doi.org/10.3115/v1/P14-2090} {Optimizing segmentation
  strategies for simultaneous speech translation}.
\newblock In \emph{Proceedings of the 52nd Annual Meeting of the Association
  for Computational Linguistics (Volume 2: Short Papers)}, pages 551--556,
  Baltimore, Maryland. Association for Computational Linguistics.

\bibitem[{Ott et~al.(2019)Ott, Edunov, Baevski, Fan, Gross, Ng, Grangier, and
  Auli}]{ott-etal-2019-fairseq}
Myle Ott, Sergey Edunov, Alexei Baevski, Angela Fan, Sam Gross, Nathan Ng,
  David Grangier, and Michael Auli. 2019.
\newblock \href {https://doi.org/10.18653/v1/N19-4009} {fairseq: A fast,
  extensible toolkit for sequence modeling}.
\newblock In \emph{Proceedings of the 2019 Conference of the North {A}merican
  Chapter of the Association for Computational Linguistics (Demonstrations)},
  pages 48--53, Minneapolis, Minnesota. Association for Computational
  Linguistics.

\bibitem[{Petek et~al.(1996)Petek, Andersen, and Dalsgaard}]{607750}
B.~Petek, O.~Andersen, and P.~Dalsgaard. 1996.
\newblock \href {https://doi.org/10.1109/ICSLP.1996.607750} {On the robust
  automatic segmentation of spontaneous speech}.
\newblock In \emph{Proceeding of Fourth International Conference on Spoken
  Language Processing. ICSLP '96}, volume~2, pages 913--916 vol.2.

\bibitem[{Pitt et~al.(2005)Pitt, Johnson, Hume, Kiesling, and
  Raymond}]{PITT200589}
Mark~A. Pitt, Keith Johnson, Elizabeth Hume, Scott Kiesling, and William
  Raymond. 2005.
\newblock \href {https://doi.org/https://doi.org/10.1016/j.specom.2004.09.001}
  {The buckeye corpus of conversational speech: labeling conventions and a test
  of transcriber reliability}.
\newblock \emph{Speech Communication}, 45(1):89--95.

\bibitem[{Popovi{\'c}(2015)}]{popovic-2015-chrf}
Maja Popovi{\'c}. 2015.
\newblock \href {https://doi.org/10.18653/v1/W15-3049} {chr{F}: character
  n-gram {F}-score for automatic {MT} evaluation}.
\newblock In \emph{Proceedings of the Tenth Workshop on Statistical Machine
  Translation}, pages 392--395, Lisbon, Portugal. Association for Computational
  Linguistics.

\bibitem[{Popovi{\'c}(2017)}]{popovic-2017-chrf}
Maja Popovi{\'c}. 2017.
\newblock \href {https://doi.org/10.18653/v1/W17-4770} {chr{F}++: words helping
  character n-grams}.
\newblock In \emph{Proceedings of the Second Conference on Machine
  Translation}, pages 612--618, Copenhagen, Denmark. Association for
  Computational Linguistics.

\bibitem[{Post(2018)}]{post-2018-call}
Matt Post. 2018.
\newblock \href {https://doi.org/10.18653/v1/W18-6319} {A call for clarity in
  reporting {BLEU} scores}.
\newblock In \emph{Proceedings of the Third Conference on Machine Translation:
  Research Papers}, pages 186--191, Brussels, Belgium. Association for
  Computational Linguistics.

\bibitem[{Raffel et~al.(2017)Raffel, Luong, Liu, Weiss, and Eck}]{LinearTime}
Colin Raffel, Minh-Thang Luong, Peter~J. Liu, Ron~J. Weiss, and Douglas Eck.
  2017.
\newblock \href {https://proceedings.mlr.press/v70/raffel17a.html} {Online and
  linear-time attention by enforcing monotonic alignments}.
\newblock In \emph{Proceedings of the 34th International Conference on Machine
  Learning}, volume~70 of \emph{Proceedings of Machine Learning Research},
  pages 2837--2846. PMLR.

\bibitem[{Rangarajan~Sridhar et~al.(2013)Rangarajan~Sridhar, Chen, Bangalore,
  Ljolje, and Chengalvarayan}]{rangarajan-sridhar-etal-2013-segmentation}
Vivek~Kumar Rangarajan~Sridhar, John Chen, Srinivas Bangalore, Andrej Ljolje,
  and Rathinavelu Chengalvarayan. 2013.
\newblock \href {https://aclanthology.org/N13-1023} {Segmentation strategies
  for streaming speech translation}.
\newblock In \emph{Proceedings of the 2013 Conference of the North {A}merican
  Chapter of the Association for Computational Linguistics: Human Language
  Technologies}, pages 230--238, Atlanta, Georgia. Association for
  Computational Linguistics.

\bibitem[{R{\"a}s{\"a}nen et~al.(2009)R{\"a}s{\"a}nen, Laine, and
  Altosaar}]{e39d9b7accfd4615ae29143a55960b0e}
Okko R{\"a}s{\"a}nen, Unto Laine, and Toomas Altosaar. 2009.
\newblock \href
  {https://citeseerx.ist.psu.edu/document?repid=rep1&type=pdf&doi=91ff68c684116aeaa4de8b407fa79bbf1e05dc3c}
  {An improved speech segmentation quality measure: the r-value}.
\newblock In \emph{10th Interspeech Conference, Brighton, UK, September 6-10,
  2009}.

\bibitem[{Ren et~al.(2020)Ren, Liu, Tan, Zhang, Qin, Zhao, and
  Liu}]{ren-etal-2020-simulspeech}
Yi~Ren, Jinglin Liu, Xu~Tan, Chen Zhang, Tao Qin, Zhou Zhao, and Tie-Yan Liu.
  2020.
\newblock \href {https://doi.org/10.18653/v1/2020.acl-main.350}
  {{S}imul{S}peech: End-to-end simultaneous speech to text translation}.
\newblock In \emph{Proceedings of the 58th Annual Meeting of the Association
  for Computational Linguistics}, pages 3787--3796, Online. Association for
  Computational Linguistics.

\bibitem[{Sakran et~al.(2017)Sakran, Abdou, Hamid, and
  Rashwan}]{sakran2017review}
Alaa~Ehab Sakran, Sherif~Mahdy Abdou, Salah~Eldeen Hamid, and Mohsen Rashwan.
  2017.
\newblock \href {https://www.academia.edu/download/53039593/V6I4201790.pdf} {A
  review: Automatic speech segmentation}.
\newblock \emph{International Journal of Computer Science and Mobile
  Computing}, 6(4):308--315.

\bibitem[{Salakhutdinov and Hinton(2009)}]{SALAKHUTDINOV2009969}
Ruslan Salakhutdinov and Geoffrey Hinton. 2009.
\newblock \href {https://doi.org/https://doi.org/10.1016/j.ijar.2008.11.006}
  {Semantic hashing}.
\newblock \emph{International Journal of Approximate Reasoning},
  50(7):969--978.
\newblock Special Section on Graphical Models and Information Retrieval.

\bibitem[{Snover et~al.(2006)Snover, Dorr, Schwartz, Micciulla, and
  Makhoul}]{snover-etal-2006-ter}
Matthew Snover, Bonnie Dorr, Rich Schwartz, Linnea Micciulla, and John Makhoul.
  2006.
\newblock \href {https://aclanthology.org/2006.amta-papers.25} {A study of
  translation edit rate with targeted human annotation}.
\newblock In \emph{Proceedings of the 7th Conference of the Association for
  Machine Translation in the Americas: Technical Papers}, pages 223--231,
  Cambridge, Massachusetts, USA. Association for Machine Translation in the
  Americas.

\bibitem[{Sohn(2016)}]{NIPS2016_6b180037}
Kihyuk Sohn. 2016.
\newblock \href
  {https://proceedings.neurips.cc/paper/2016/file/6b180037abbebea991d8b1232f8a8ca9-Paper.pdf}
  {Improved deep metric learning with multi-class n-pair loss objective}.
\newblock In \emph{Advances in Neural Information Processing Systems},
  volume~29. Curran Associates, Inc.

\bibitem[{Tang et~al.(2021{\natexlab{a}})Tang, Pino, Li, Wang, and
  Genzel}]{tang-etal-2021-improving}
Yun Tang, Juan Pino, Xian Li, Changhan Wang, and Dmitriy Genzel.
  2021{\natexlab{a}}.
\newblock \href {https://doi.org/10.18653/v1/2021.acl-long.328} {Improving
  speech translation by understanding and learning from the auxiliary text
  translation task}.
\newblock In \emph{Proceedings of the 59th Annual Meeting of the Association
  for Computational Linguistics and the 11th International Joint Conference on
  Natural Language Processing (Volume 1: Long Papers)}, pages 4252--4261,
  Online. Association for Computational Linguistics.

\bibitem[{Tang et~al.(2021{\natexlab{b}})Tang, Pino, Wang, Ma, and
  Genzel}]{9415058}
Yun Tang, Juan Pino, Changhan Wang, Xutai Ma, and Dmitriy Genzel.
  2021{\natexlab{b}}.
\newblock \href {https://doi.org/10.1109/ICASSP39728.2021.9415058} {A general
  multi-task learning framework to leverage text data for speech to text
  tasks}.
\newblock In \emph{ICASSP 2021 - 2021 IEEE International Conference on
  Acoustics, Speech and Signal Processing (ICASSP)}, pages 6209--6213.

\bibitem[{Valentini-Botinhao and King(2021)}]{valentini2021detection}
Cassia Valentini-Botinhao and Simon King. 2021.
\newblock \href
  {https://www.research.ed.ac.uk/en/publications/detection-and-analysis-of-attention-errors-in-sequence-to-sequenc}
  {Detection and analysis of attention errors in sequence-to-sequence
  text-to-speech}.
\newblock In \emph{Interspeech 2021: The 22nd Annual Conference of the
  International Speech Communication Association}, pages 2746--2750. ISCA.

\bibitem[{Vaswani et~al.(2017)Vaswani, Shazeer, Parmar, Uszkoreit, Jones,
  Gomez, Kaiser, and Polosukhin}]{NIPS2017_7181}
Ashish Vaswani, Noam Shazeer, Niki Parmar, Jakob Uszkoreit, Llion Jones,
  Aidan~N Gomez, \L~ukasz Kaiser, and Illia Polosukhin. 2017.
\newblock \href
  {http://papers.nips.cc/paper/7181-attention-is-all-you-need.pdf} {Attention
  is all you need}.
\newblock In I.~Guyon, U.~V. Luxburg, S.~Bengio, H.~Wallach, R.~Fergus,
  S.~Vishwanathan, and R.~Garnett, editors, \emph{Advances in Neural
  Information Processing Systems 30}, pages 5998--6008. Curran Associates, Inc.

\bibitem[{Vig and Belinkov(2019)}]{vig-belinkov-2019-analyzing}
Jesse Vig and Yonatan Belinkov. 2019.
\newblock \href {https://doi.org/10.18653/v1/W19-4808} {Analyzing the structure
  of attention in a transformer language model}.
\newblock In \emph{Proceedings of the 2019 ACL Workshop BlackboxNLP: Analyzing
  and Interpreting Neural Networks for NLP}, pages 63--76, Florence, Italy.
  Association for Computational Linguistics.

\bibitem[{Wang et~al.(2020)Wang, Tang, Ma, Wu, Okhonko, and
  Pino}]{wang-etal-2020-fairseq}
Changhan Wang, Yun Tang, Xutai Ma, Anne Wu, Dmytro Okhonko, and Juan Pino.
  2020.
\newblock \href {https://aclanthology.org/2020.aacl-demo.6} {Fairseq {S}2{T}:
  Fast speech-to-text modeling with fairseq}.
\newblock In \emph{Proceedings of the 1st Conference of the Asia-Pacific
  Chapter of the Association for Computational Linguistics and the 10th
  International Joint Conference on Natural Language Processing: System
  Demonstrations}, pages 33--39, Suzhou, China. Association for Computational
  Linguistics.

\bibitem[{Wang and Liu(2021)}]{Wang_2021_CVPR}
Feng Wang and Huaping Liu. 2021.
\newblock \href
  {https://openaccess.thecvf.com/content/CVPR2021/papers/Wang_Understanding_the_Behaviour_of_Contrastive_Loss_CVPR_2021_paper.pdf}
  {Understanding the behaviour of contrastive loss}.
\newblock In \emph{Proceedings of the IEEE/CVF Conference on Computer Vision
  and Pattern Recognition (CVPR)}, pages 2495--2504.

\bibitem[{Weiss et~al.(2017)Weiss, Chorowski, Jaitly, Wu, and Chen}]{Weiss2017}
Ron~J. Weiss, Jan Chorowski, Navdeep Jaitly, Yonghui Wu, and Zhifeng Chen.
  2017.
\newblock \href {https://doi.org/10.21437/Interspeech.2017-503}
  {Sequence-to-sequence models can directly translate foreign speech}.
\newblock In \emph{Proc. Interspeech 2017}, pages 2625--2629.

\bibitem[{Wu et~al.(2021)Wu, Wang, Xia, Liu, Wu, Xie, Qin, and
  Liu}]{pmlr-v139-wu21e}
Xueqing Wu, Lewen Wang, Yingce Xia, Weiqing Liu, Lijun Wu, Shufang Xie, Tao
  Qin, and Tie-Yan Liu. 2021.
\newblock \href {https://proceedings.mlr.press/v139/wu21e.html} {Temporally
  correlated task scheduling for sequence learning}.
\newblock In \emph{Proceedings of the 38th International Conference on Machine
  Learning}, volume 139 of \emph{Proceedings of Machine Learning Research},
  pages 11274--11284. PMLR.

\bibitem[{Yang et~al.(2018)Yang, Tu, Wong, Meng, Chao, and
  Zhang}]{yang-etal-2018-modeling}
Baosong Yang, Zhaopeng Tu, Derek~F. Wong, Fandong Meng, Lidia~S. Chao, and Tong
  Zhang. 2018.
\newblock \href {https://doi.org/10.18653/v1/D18-1475} {Modeling localness for
  self-attention networks}.
\newblock In \emph{Proceedings of the 2018 Conference on Empirical Methods in
  Natural Language Processing}, pages 4449--4458, Brussels, Belgium.
  Association for Computational Linguistics.

\bibitem[{Yang et~al.(2020)Yang, Lu, Kang, Xue, Xiao, Su, Xie, and
  Yu}]{YANG2020121}
Shan Yang, Heng Lu, Shiyin Kang, Liumeng Xue, Jinba Xiao, Dan Su, Lei Xie, and
  Dong Yu. 2020.
\newblock \href {https://doi.org/https://doi.org/10.1016/j.neunet.2020.01.034}
  {On the localness modeling for the self-attention based end-to-end speech
  synthesis}.
\newblock \emph{Neural Networks}, 125:121--130.

\bibitem[{Yarmohammadi et~al.(2013)Yarmohammadi, Rangarajan~Sridhar, Bangalore,
  and Sankaran}]{yarmohammadi-etal-2013-incremental}
Mahsa Yarmohammadi, Vivek~Kumar Rangarajan~Sridhar, Srinivas Bangalore, and
  Baskaran Sankaran. 2013.
\newblock \href {https://aclanthology.org/I13-1141} {Incremental segmentation
  and decoding strategies for simultaneous translation}.
\newblock In \emph{Proceedings of the Sixth International Joint Conference on
  Natural Language Processing}, pages 1032--1036, Nagoya, Japan. Asian
  Federation of Natural Language Processing.

\bibitem[{Ye et~al.(2022)Ye, Wang, and Li}]{ye-etal-2022-cross}
Rong Ye, Mingxuan Wang, and Lei Li. 2022.
\newblock \href {https://doi.org/10.18653/v1/2022.naacl-main.376} {Cross-modal
  contrastive learning for speech translation}.
\newblock In \emph{Proceedings of the 2022 Conference of the North American
  Chapter of the Association for Computational Linguistics: Human Language
  Technologies}, pages 5099--5113, Seattle, United States. Association for
  Computational Linguistics.

\bibitem[{Zaidi et~al.(2022)Zaidi, Lee, Kim, and Kim}]{zaidi22_interspeech}
Mohd~Abbas Zaidi, Beomseok Lee, Sangha Kim, and Chanwoo Kim. 2022.
\newblock \href {https://doi.org/10.21437/Interspeech.2022-10617} {{Cross-Modal
  Decision Regularization for Simultaneous Speech Translation}}.
\newblock In \emph{Proc. Interspeech 2022}, pages 116--120.

\bibitem[{Zeng et~al.(2021)Zeng, Li, and Liu}]{zeng-etal-2021-realtrans}
Xingshan Zeng, Liangyou Li, and Qun Liu. 2021.
\newblock \href {https://doi.org/10.18653/v1/2021.findings-acl.218}
  {{R}eal{T}ran{S}: End-to-end simultaneous speech translation with
  convolutional weighted-shrinking transformer}.
\newblock In \emph{Findings of the Association for Computational Linguistics:
  ACL-IJCNLP 2021}, pages 2461--2474, Online. Association for Computational
  Linguistics.

\bibitem[{Zhang et~al.(2022{\natexlab{a}})Zhang, He, Wu, and
  Wang}]{zhang-etal-2022-learning}
Ruiqing Zhang, Zhongjun He, Hua Wu, and Haifeng Wang. 2022{\natexlab{a}}.
\newblock \href {https://doi.org/10.18653/v1/2022.acl-long.542} {Learning
  adaptive segmentation policy for end-to-end simultaneous translation}.
\newblock In \emph{Proceedings of the 60th Annual Meeting of the Association
  for Computational Linguistics (Volume 1: Long Papers)}, pages 7862--7874,
  Dublin, Ireland. Association for Computational Linguistics.

\bibitem[{Zhang and Feng(2021{\natexlab{a}})}]{zhang-feng-2021-icts}
Shaolei Zhang and Yang Feng. 2021{\natexlab{a}}.
\newblock \href {https://doi.org/10.18653/v1/2021.autosimtrans-1.1} {{ICT}{'}s
  system for {A}uto{S}im{T}rans 2021: Robust char-level simultaneous
  translation}.
\newblock In \emph{Proceedings of the Second Workshop on Automatic Simultaneous
  Translation}, pages 1--11, Online. Association for Computational Linguistics.

\bibitem[{Zhang and
  Feng(2021{\natexlab{b}})}]{zhang-feng-2021-modeling-concentrated}
Shaolei Zhang and Yang Feng. 2021{\natexlab{b}}.
\newblock \href {https://doi.org/10.18653/v1/2021.findings-emnlp.121} {Modeling
  concentrated cross-attention for neural machine translation with {G}aussian
  mixture model}.
\newblock In \emph{Findings of the Association for Computational Linguistics:
  EMNLP 2021}, pages 1401--1411, Punta Cana, Dominican Republic. Association
  for Computational Linguistics.

\bibitem[{Zhang and Feng(2021{\natexlab{c}})}]{zhang-feng-2021-universal}
Shaolei Zhang and Yang Feng. 2021{\natexlab{c}}.
\newblock \href {https://doi.org/10.18653/v1/2021.emnlp-main.581} {Universal
  simultaneous machine translation with mixture-of-experts wait-k policy}.
\newblock In \emph{Proceedings of the 2021 Conference on Empirical Methods in
  Natural Language Processing}, pages 7306--7317, Online and Punta Cana,
  Dominican Republic. Association for Computational Linguistics.

\bibitem[{Zhang and Feng(2022{\natexlab{a}})}]{gma}
Shaolei Zhang and Yang Feng. 2022{\natexlab{a}}.
\newblock \href {https://doi.org/10.18653/v1/2022.findings-acl.238} {{G}aussian
  multi-head attention for simultaneous machine translation}.
\newblock In \emph{Findings of the Association for Computational Linguistics:
  ACL 2022}, pages 3019--3030, Dublin, Ireland. Association for Computational
  Linguistics.

\bibitem[{Zhang and Feng(2022{\natexlab{b}})}]{ITST}
Shaolei Zhang and Yang Feng. 2022{\natexlab{b}}.
\newblock \href {https://aclanthology.org/2022.emnlp-main.65}
  {Information-transport-based policy for simultaneous translation}.
\newblock In \emph{Proceedings of the 2022 Conference on Empirical Methods in
  Natural Language Processing}, pages 992--1013, Abu Dhabi, United Arab
  Emirates. Association for Computational Linguistics.

\bibitem[{Zhang and Feng(2022{\natexlab{c}})}]{dualpath}
Shaolei Zhang and Yang Feng. 2022{\natexlab{c}}.
\newblock \href {https://doi.org/10.18653/v1/2022.acl-long.176} {Modeling dual
  read/write paths for simultaneous machine translation}.
\newblock In \emph{Proceedings of the 60th Annual Meeting of the Association
  for Computational Linguistics (Volume 1: Long Papers)}, pages 2461--2477,
  Dublin, Ireland. Association for Computational Linguistics.

\bibitem[{Zhang and Feng(2022{\natexlab{d}})}]{laf}
Shaolei Zhang and Yang Feng. 2022{\natexlab{d}}.
\newblock \href {https://doi.org/10.18653/v1/2022.acl-long.467} {Reducing
  position bias in simultaneous machine translation with length-aware
  framework}.
\newblock In \emph{Proceedings of the 60th Annual Meeting of the Association
  for Computational Linguistics (Volume 1: Long Papers)}, pages 6775--6788,
  Dublin, Ireland. Association for Computational Linguistics.

\bibitem[{Zhang and Feng(2023)}]{zhang2023hidden}
Shaolei Zhang and Yang Feng. 2023.
\newblock \href {https://openreview.net/forum?id=9y0HFvaAYD6} {Hidden markov
  transformer for simultaneous machine translation}.
\newblock In \emph{The Eleventh International Conference on Learning
  Representations}.

\bibitem[{Zhang et~al.(2021)Zhang, Feng, and Li}]{future-guided}
Shaolei Zhang, Yang Feng, and Liangyou Li. 2021.
\newblock \href {https://ojs.aaai.org/index.php/AAAI/article/view/17696}
  {Future-guided incremental transformer for simultaneous translation}.
\newblock \emph{Proceedings of the AAAI Conference on Artificial Intelligence},
  35(16):14428--14436.

\bibitem[{Zhang et~al.(2022{\natexlab{b}})Zhang, Guo, and Feng}]{wait-info}
Shaolei Zhang, Shoutao Guo, and Yang Feng. 2022{\natexlab{b}}.
\newblock \href {https://aclanthology.org/2022.findings-emnlp.166} {Wait-info
  policy: Balancing source and target at information level for simultaneous
  machine translation}.
\newblock In \emph{Findings of the Association for Computational Linguistics:
  EMNLP 2022}, pages 2249--2263, Abu Dhabi, United Arab Emirates. Association
  for Computational Linguistics.

\bibitem[{Zheng et~al.(2020)Zheng, Li, Xie, and Lu}]{9054148}
Yibin Zheng, Xinhui Li, Fenglong Xie, and Li~Lu. 2020.
\newblock \href {https://doi.org/10.1109/ICASSP40776.2020.9054148} {Improving
  end-to-end speech synthesis with local recurrent neural network enhanced
  transformer}.
\newblock In \emph{ICASSP 2020 - 2020 IEEE International Conference on
  Acoustics, Speech and Signal Processing (ICASSP)}, pages 6734--6738.

\end{thebibliography}
\bibliographystyle{acl_natbib}
\clearpage

\appendix

\section{Dynamic Programming for Expected Feature-to-Segment Mapping}
\label{sec:dp}

In Sec.\ref{sec:training}, we propose the expected feature-to-segment map to get the expected segment representations while keeping segment representations differentiable to the segmentation. Expected feature-to-segment map calculates the probability $p\!\left ( a_{i} \in \mathrm{Seg}_{k} \right )$ that the speech feature $a_{i}$ belongs to the $k^{th}$ segment $\mathrm{Seg}_{k}$, and then gets the expected segment representation by weighting all speech features $a_{i}$ with $p\!\left ( a_{i} \in \mathrm{Seg}_{k} \right )$.

Given segmentation probability $p_{i}$, we calculate $p\!\left ( a_{i} \in \mathrm{Seg}_{k} \right )$ via dynamic programming. Whether $a_{i}$ belongs to the $k^{th}$ segment depends on which segment that speech feature $a_{i-1}$ is located in, consisting of 3 situations:
\begin{itemize}
    \item $a_{i-1} \in \mathrm{Seg}_{k-1}$: If DiSeg segments at feature $a_{i-1}$ (with probability $p_{i-1}$), then $a_{i}$ belongs to $\mathrm{Seg}_{k}$;
    \item $a_{i-1} \in \mathrm{Seg}_{k}$: If DiSeg does not segment at feature $a_{i-1}$ (with probability $1-p_{i-1}$), then $a_{i}$ belongs to $\mathrm{Seg}_{k}$;
    \item Others: $a_{i}$ can not belong to $\mathrm{Seg}_{k}$ anyway, because the feature-to-segment mapping must be monotonic.
\end{itemize}
By combining these situations, $p\!\left ( a_{i} \in \mathrm{Seg}_{k} \right )$ is calculated as:
\begin{gather}
\begin{aligned}
    p\!\left ( a_{i} \in \mathrm{Seg}_{k} \right )= p\!\left (a_{i-1} \in \mathrm{Seg}_{k-1}  \right )\times p_{i-1}& \\
    +\; p\!\left (a_{i-1} \in \mathrm{Seg}_{k}  \right )\times \left (1- p_{i-1} \right ) &.
\end{aligned}
\end{gather}
For the initialization, $p\!\left ( a_{1} \in \mathrm{Seg}_{k} \right )$ is calculated as:
\begin{gather}
    p\!\left ( a_{1} \in \mathrm{Seg}_{k} \right )=\begin{cases}
1 & \text{ if } k=1 \\
0 & \text{ if } k\neq1 
\end{cases},
\end{gather}
where the first feature inevitably belongs to the first segment. Since we constrained the number of segments to be $K$ during training, we truncate $p\!\left ( a_{i} \in \mathrm{Seg}_{k} \right )$ at $p\!\left ( a_{i} \in \mathrm{Seg}_{K} \right )$.

\section{Metrics of Word Segmentation Task}
\label{sec:Metrics of Word Segmentation Task}
In Sec.\ref{sec:Segmentation Quality}, we evaluate the segmentation quality of DiSeg on the automatic speech segmentation task \citep{10.1007/3-540-46154-X_38,sakran2017review}, and here we give the specific calculation of metrics for the speech segmentation task.

Precision (P), recall (R) and the corresponding F1 score are used to measure whether the segmentation position is correct compared with the ground-truth segmentation. Over-segmentation (OS) \citep{607750} measures whether the number of segments generated by the model is accurate, calculated as:
\begin{gather}
    \textrm{OS}=\frac{R}{P}-1,
\end{gather}
where $\textrm{OS}=0$ means that the number of segments is completely accurate, a larger OS means more segments, and a smaller OS means fewer segments. Since a large number of segments is easy to obtain a high recall score while a poor OS score, a more robust metric R-value \citep{e39d9b7accfd4615ae29143a55960b0e} is proposed to measure recall score and OS score together. R-value is calculated as:
\begin{align}
    \textrm{R-value}=&\;1-\frac{\left|r_{1} \right|+\left|r_{2} \right|}{2}, \\
    \text{where}\;\;r_{1}=&\;\sqrt{\left ( 1-R \right )^{2}+\mathrm{OS}^{2}}, \\
    r_{2}=&\;\frac{-\textrm{OS}+R-1}{\sqrt{2}}.
\end{align}
A larger $\textrm{R-value}$ indicates better segmentation quality, and the only way to achieve a perfect $\textrm{R-value}$ is to get a perfect recall score (i.e., $R=1$) and a perfect OS score (i.e., $\textrm{OS}=0$).

\section{Extended Analyses}
\label{sec:Extended Analyses}

\subsection{Effectiveness of Contrastive Learning}

\begin{figure}[t]
    \centering
    \includegraphics[width=3in]{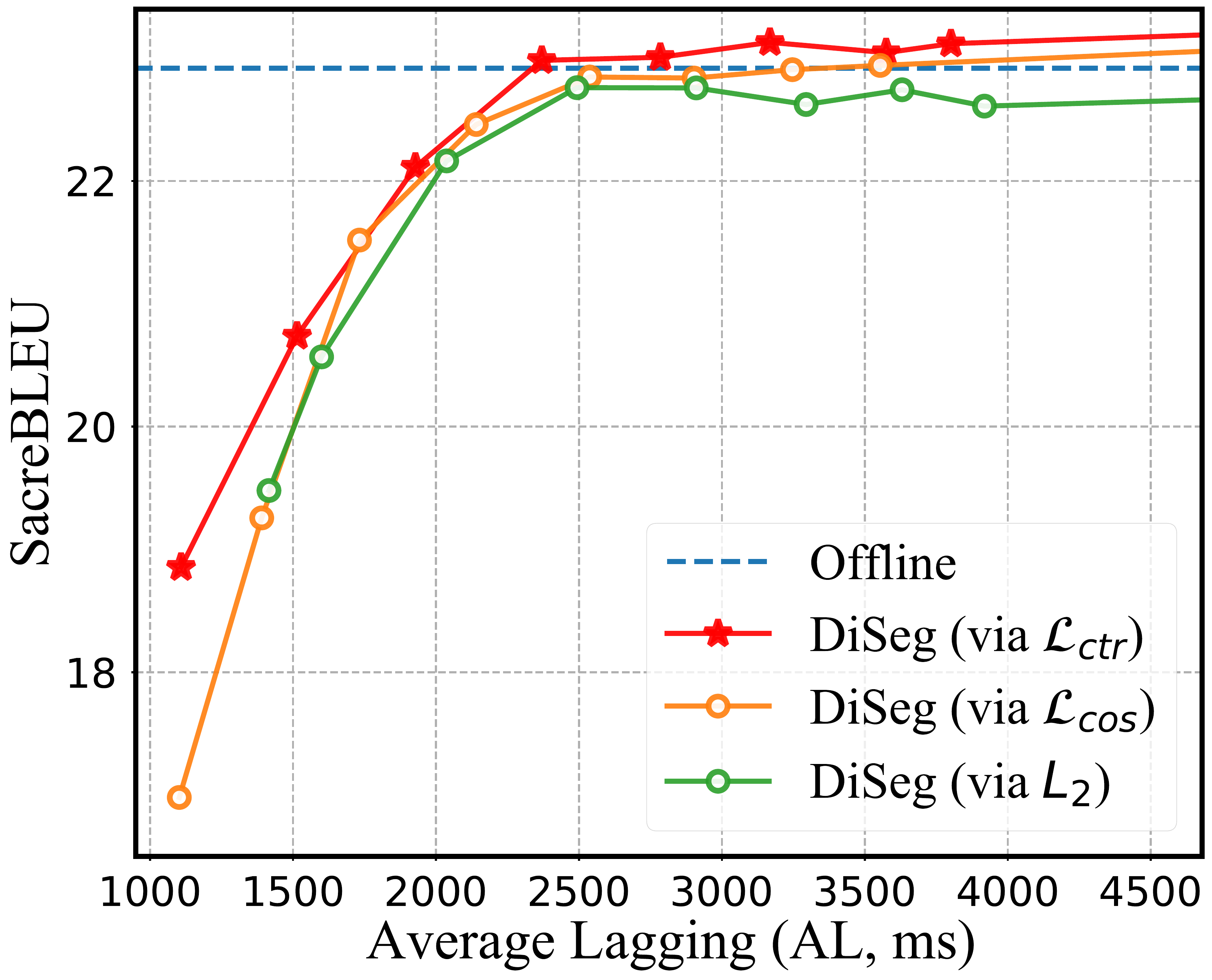}
    \caption{Comparison of semantic supervision with contrastive learning loss $\mathcal{L}_{ctr}$ and $L_{2}$ loss.}
    \label{fig:ctr_l2}
\end{figure}

During expectation training, we use the word representations to supervise the segment representations via contrastive learning $\mathcal{L}_{ctr}$ \citep{NIPS2016_6b180037}. To verify its effect, we compared the performance of applying contrastive learning loss $\mathcal{L}_{ctr}$ and some other loss functions for semantic supervision in Figure \ref{fig:ctr_l2}, including
\begin{itemize}
    \item $\mathcal{L}_{ctr}$: reduce the cosine similarity between the expected segment representation $f^{s}_{k}$ and the corresponding word representation $f^{t}_{k}$ and meanwhile separates $f^{s}_{k}$ with the rest of the word representations;
    \item $\mathcal{L}_{cos}$: reduce the cosine similarity between the expected segment representation $f^{s}_{k}$ and the corresponding word representation $f^{t}_{k}$.
    \item $L_{2}$: reduce the $L_{2}$ distance between the expected segment representation $f^{s}_{k}$ and the corresponding word representation $f^{t}_{k}$.
\end{itemize}

The results show that the cosine similarity $\mathcal{L}_{cos}$ is better than the $L_{2}$ distance to measure the difference between the segment representation and the word representation \citep{9226466,ye-etal-2022-cross}, and the contrastive learning $\mathcal{L}_{ctr}$ further improves DiSeg performance by introducing negative examples.
In particular, since $\mathcal{L}_{cos}$ and $L_{2}$ loss fails to separate the representation of the segment and non-corresponding words, it is easy to cause the segment corresponds to more words or fewer words, which can still reduce $\mathcal{L}_{cos}$ or $L_{2}$ loss but is not conducive to learning the precise segmentation. By making positive pairs (i.e., segment and the corresponding word) attracted and negative pairs (i.e., segment and those non-corresponding words) separated \citep{NIPS2016_6b180037}, contrastive learning $\mathcal{L}_{ctr}$ can learn more precise segmentation boundaries and thereby achieve better performance.

\subsection{Visualization of Segmented Attention}
\label{sec:Visualization of Segmented Attention}

\begin{figure*}[p]
\centering
\subfigure[Uni-directional attention.]{
\includegraphics[width=2in]{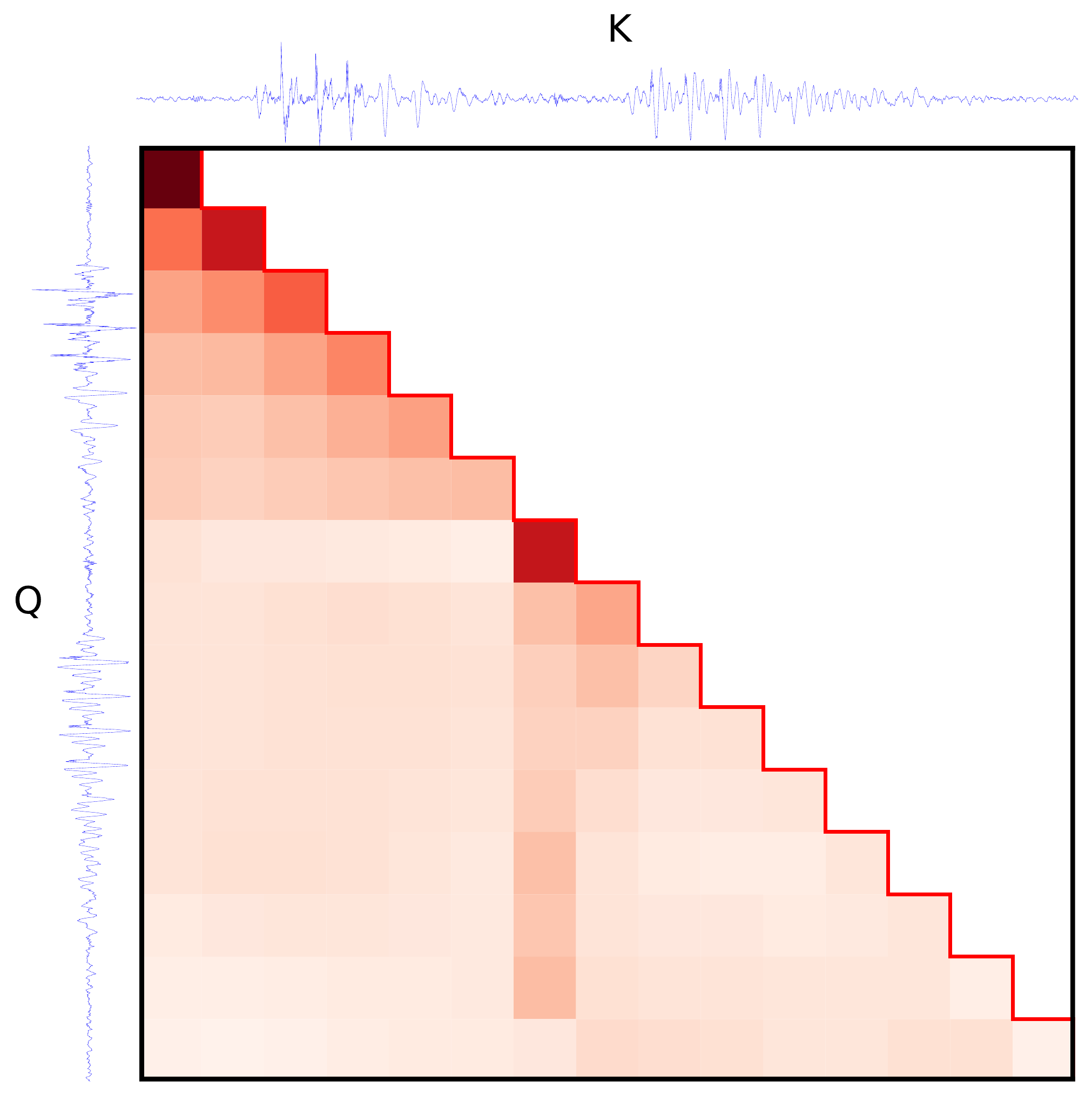}\label{fig:attn_vis11}
}
\subfigure[Segmented attention.]{
\includegraphics[width=2in]{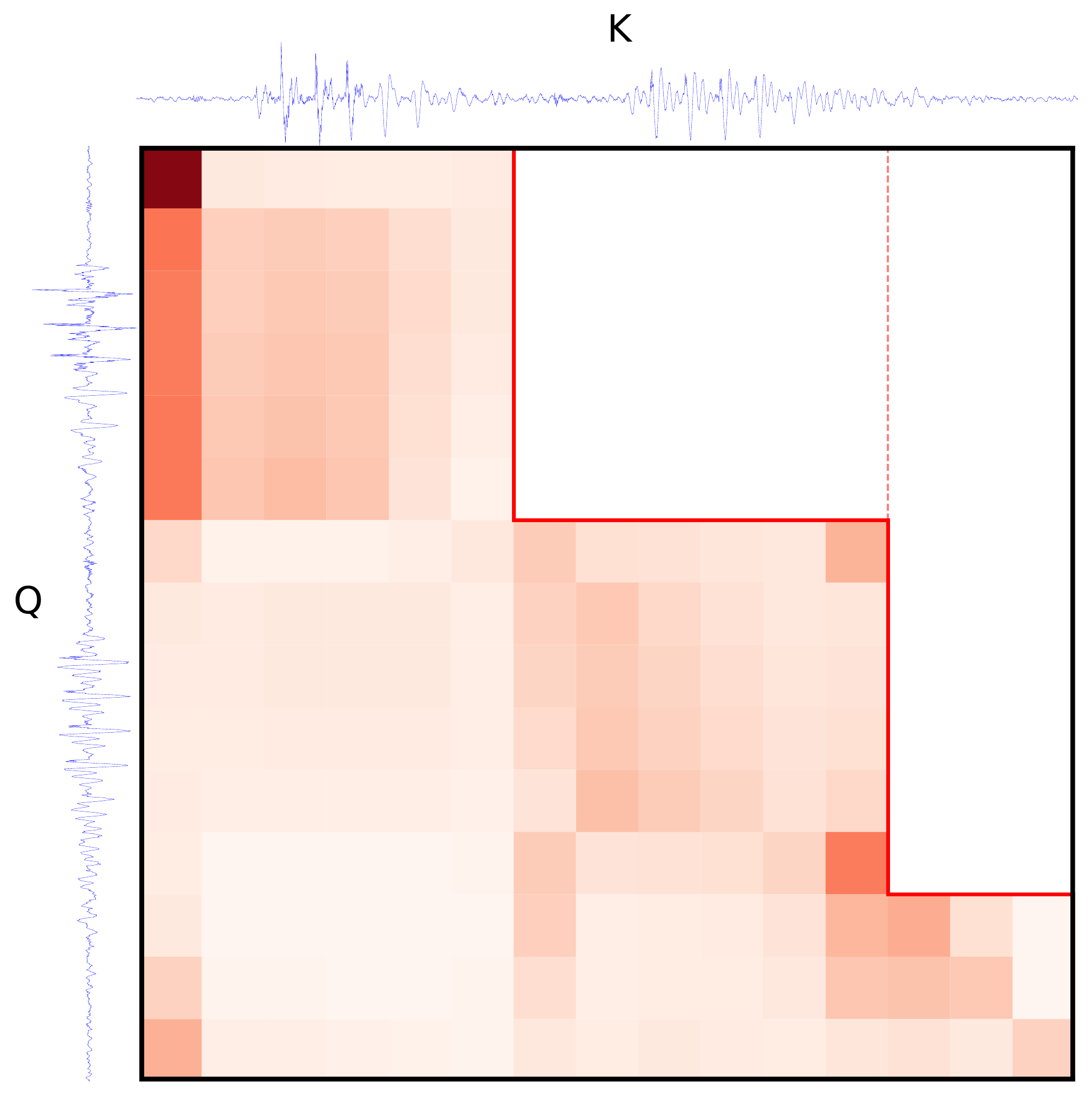}\label{fig:attn_vis12}
}
\subfigure[Bi-directional attention.]{
\includegraphics[width=2in]{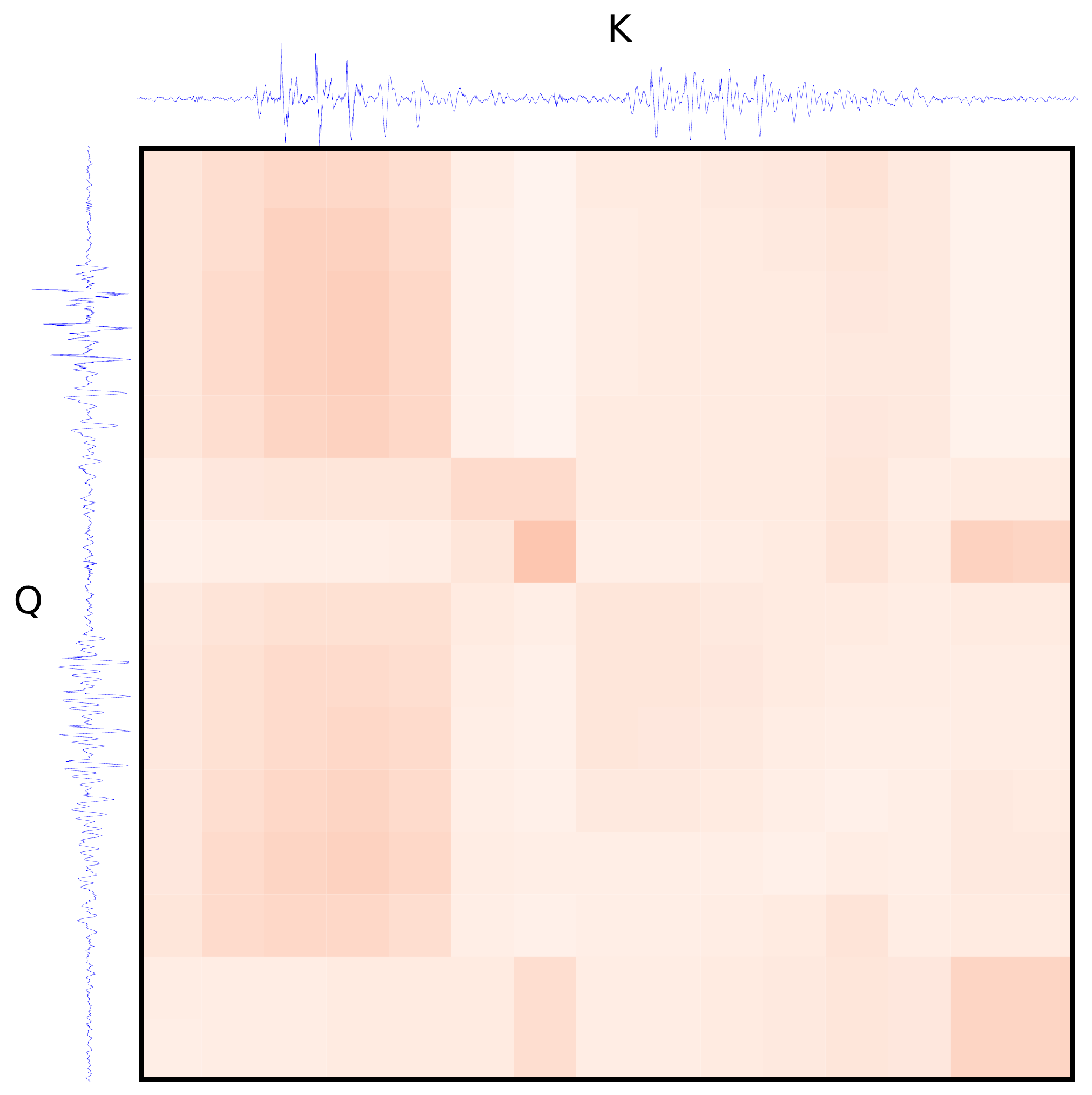}\label{fig:attn_vis13}
}
\caption{Visualization of segmented attention and uni-/bi-directional attention on short speech Ted\_1160\_70. Duration: 0.31$s$; Transcription: `\textit{Thank you.}'. Three speech segments in the segmented attention correspond to `\textit{Thank}', `\textit{you}' and silence in the transcription respectively. The color shade indicates the attention weight, and the blank area indicates that the attention is masked out.}
\label{fig:attn_vis1}
\end{figure*}

\begin{figure*}[p]
\centering
\subfigure[Uni-directional attention.]{
\includegraphics[width=2in]{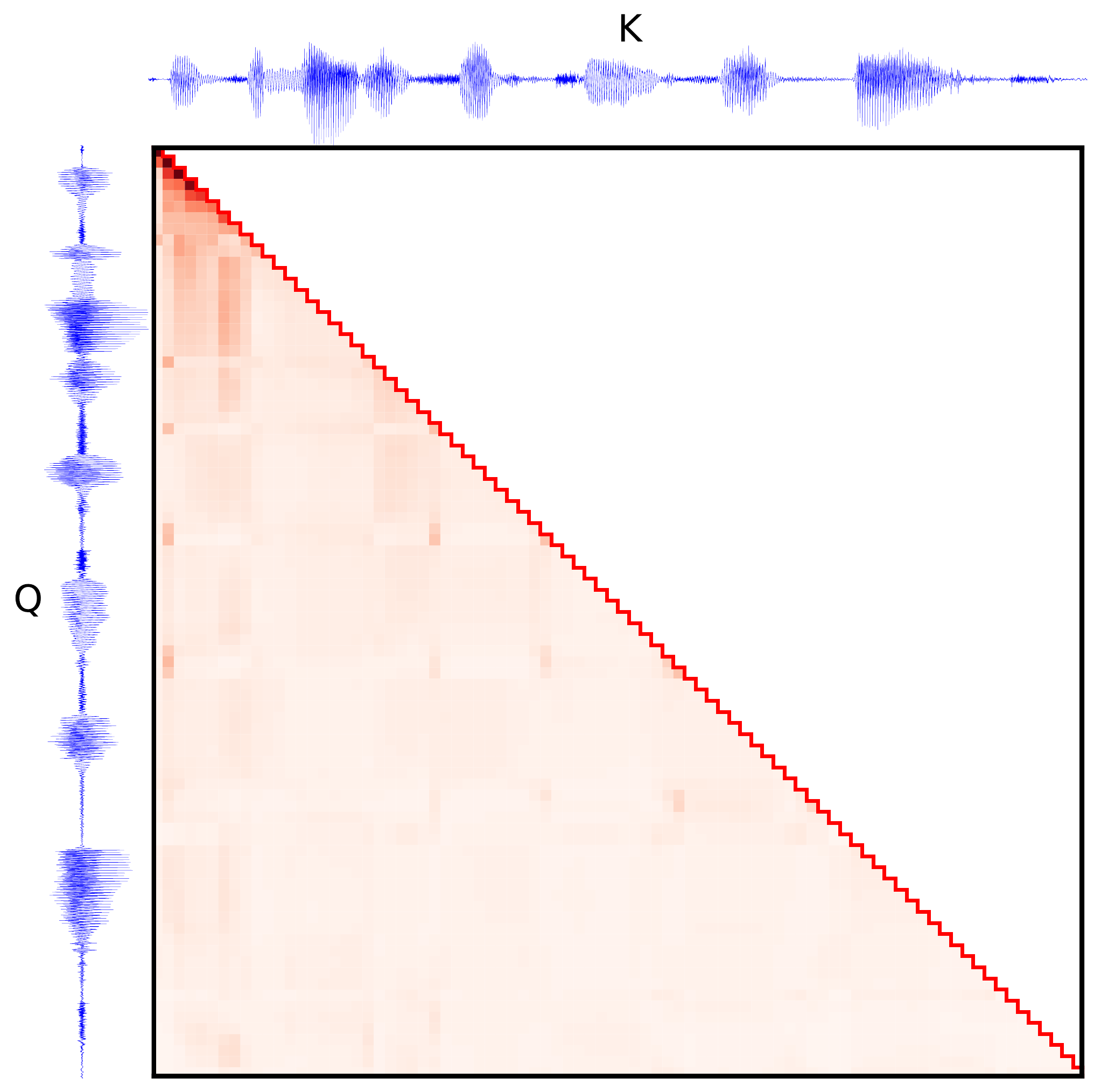}\label{fig:attn_vis21}
}
\subfigure[Segmented attention.]{
\includegraphics[width=2in]{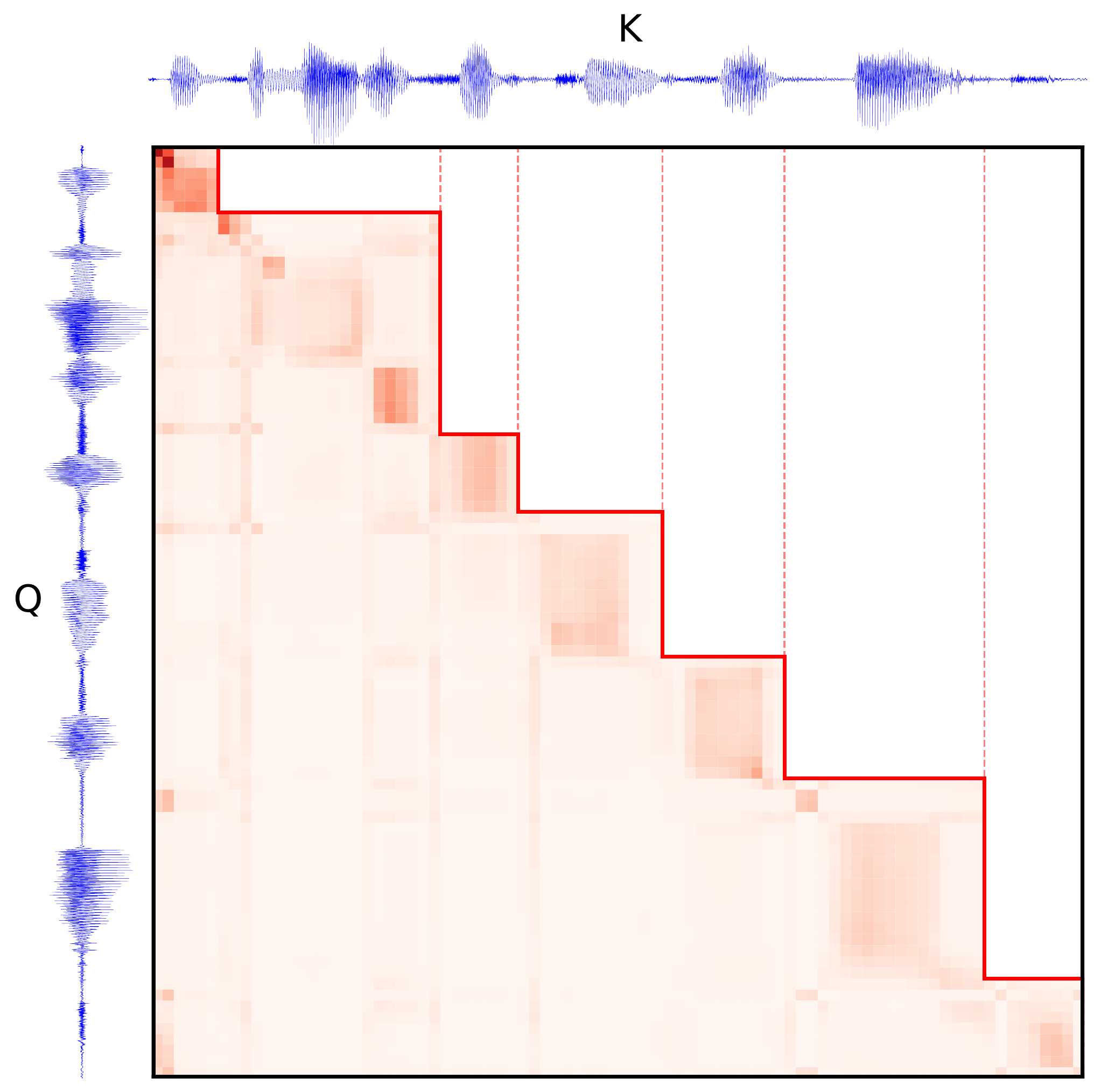}\label{fig:attn_vis22}
}
\subfigure[Bi-directional attention.]{
\includegraphics[width=2in]{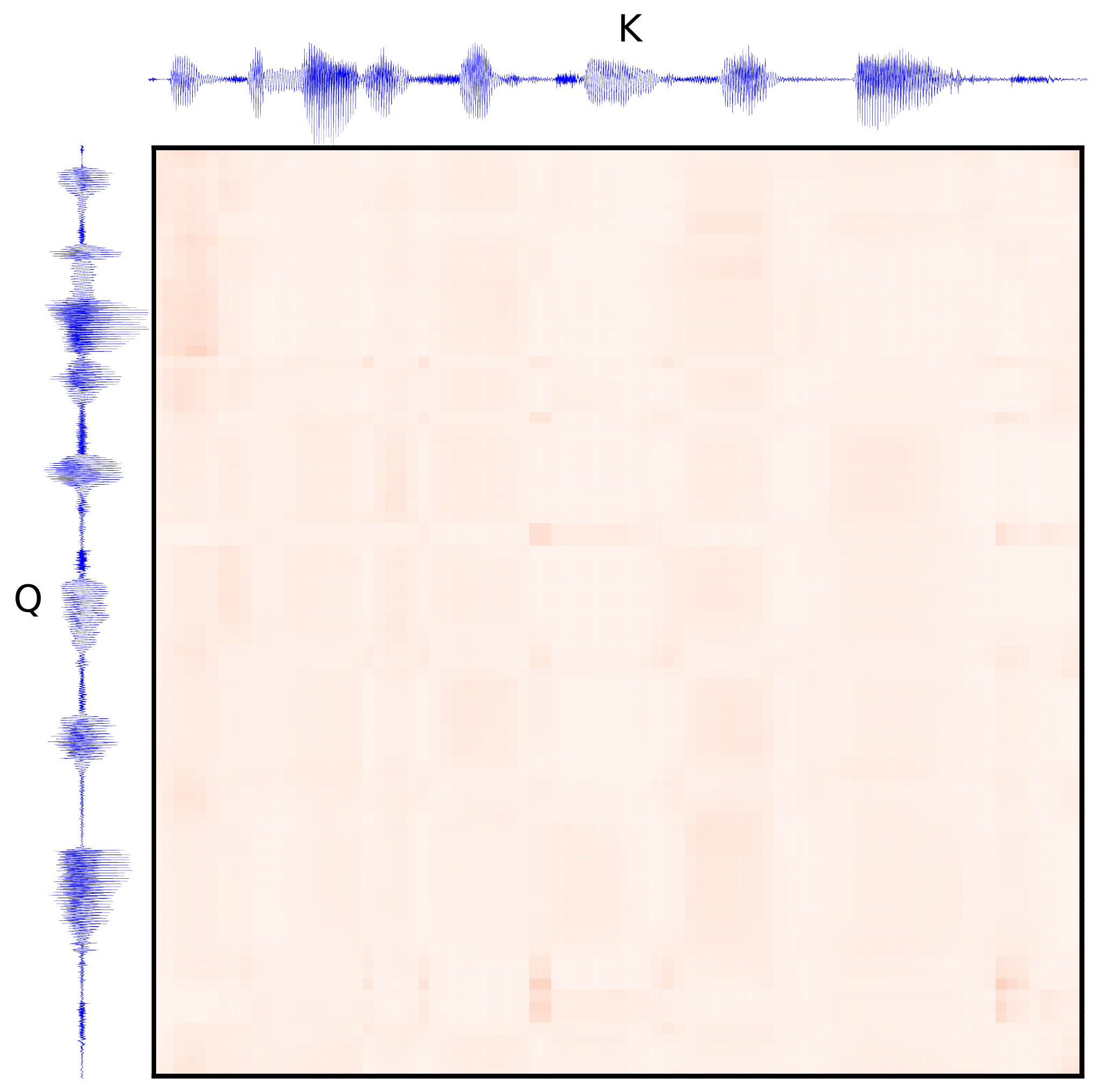}\label{fig:attn_vis23}
}
\caption{Visualization of segmented attention and uni-/bi-directional attention on medium length speech Ted\_1171\_11. Duration: 1.69$s$; Transcription: `\textit{That's about a 15-foot boat.}'. Seven speech segments in the segmented attention correspond to `\textit{That's}', `\textit{about}', `\textit{a}', `\textit{15}', `\textit{foot}', `\textit{boat}', and silence in the transcription respectively. The color shade indicates the attention weight, and the blank area indicates that the attention is masked out.}
\label{fig:attn_vis2}
\end{figure*}

\begin{figure*}[p]
\centering
\subfigure[Uni-directional attention.]{
\includegraphics[width=2in]{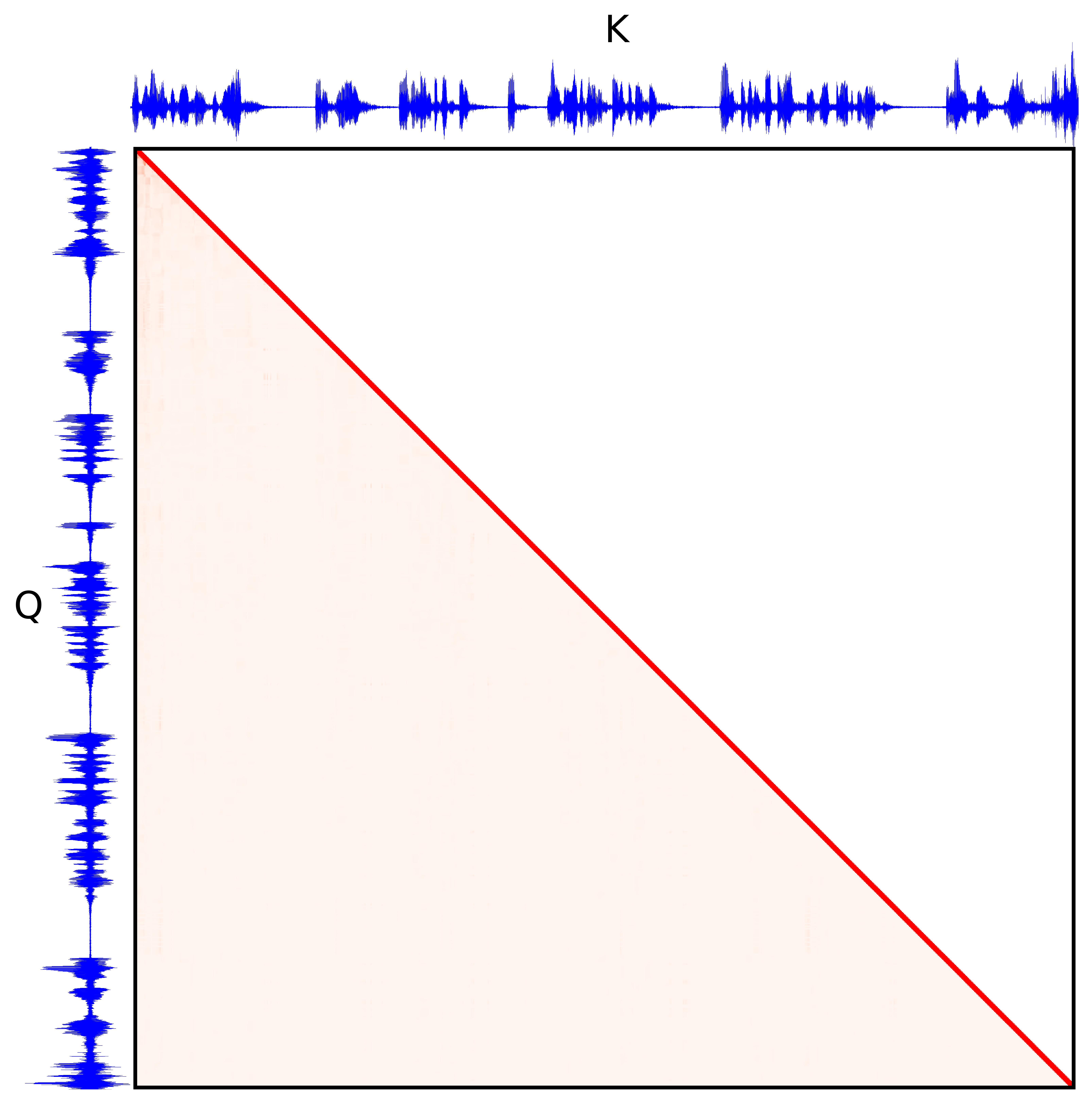}\label{fig:attn_vis31}
}
\subfigure[Segmented attention.]{
\includegraphics[width=2in]{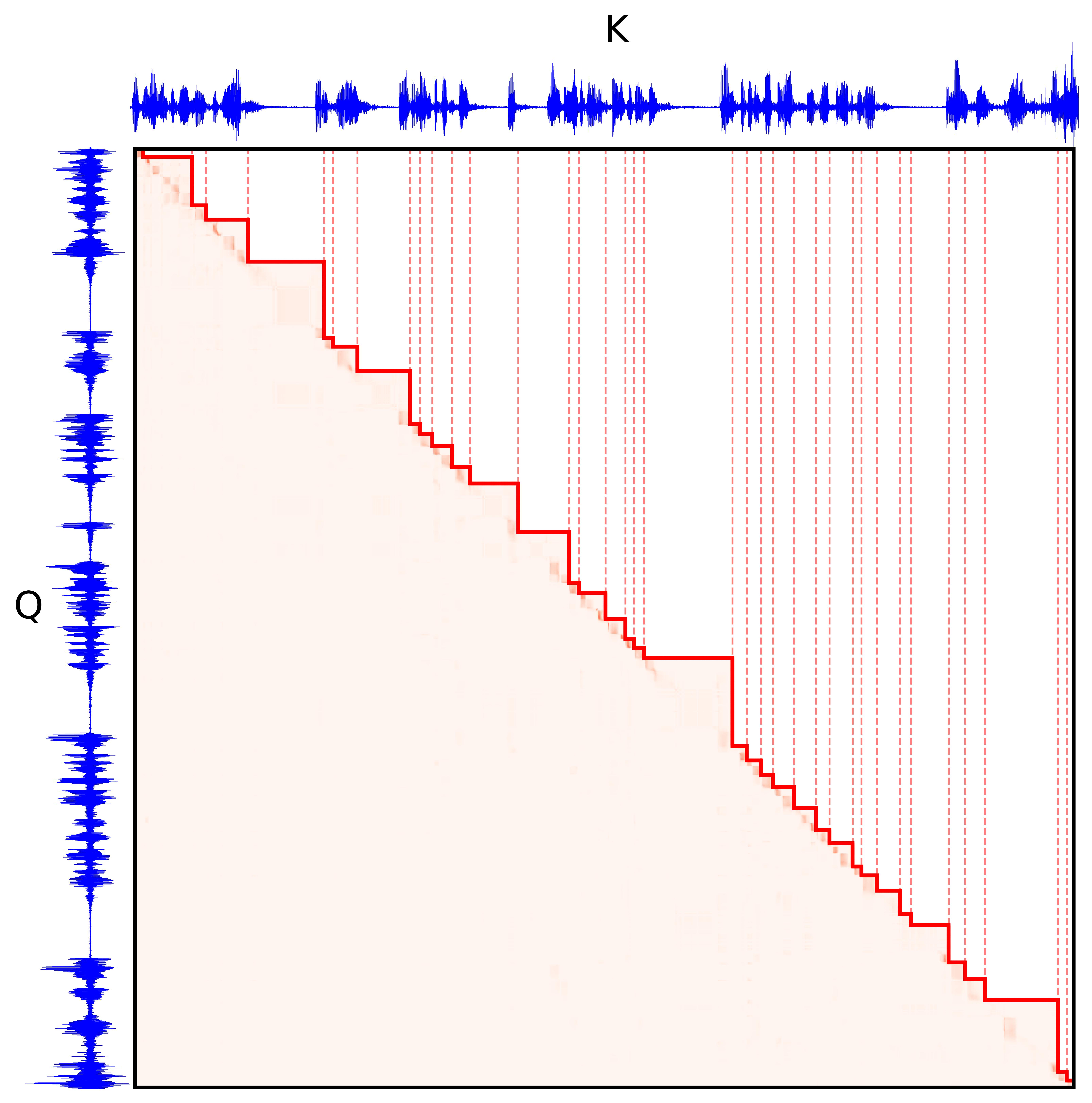}\label{fig:attn_vis32}
}
\subfigure[Bi-directional attention.]{
\includegraphics[width=2in]{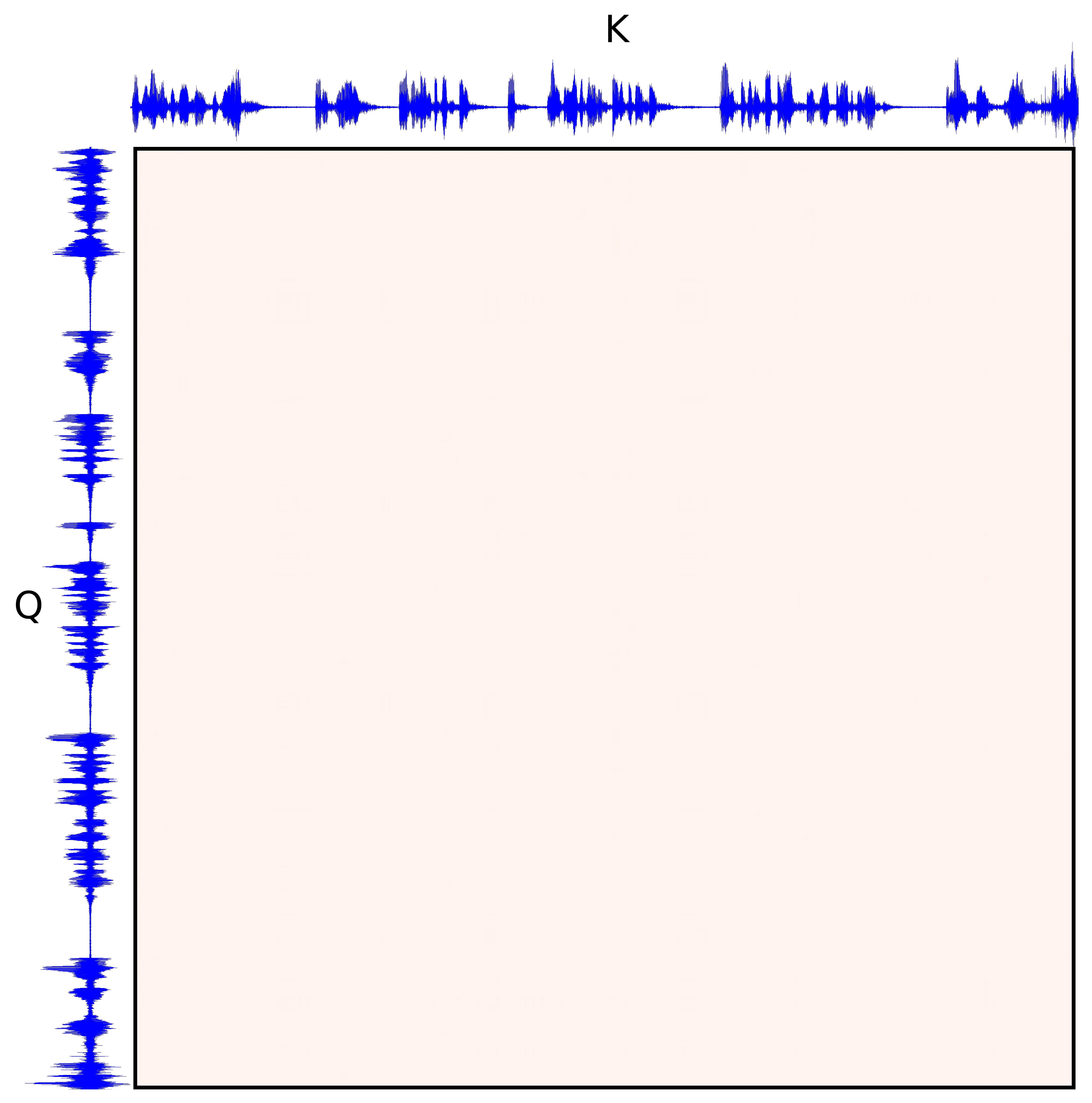}\label{fig:attn_vis33}
}
\caption{Visualization of segmented attention and uni-/bi-directional attention on extremely long speech Ted\_1104\_16. Duration: 17.09$s$; Transcription: `\textit{Let me now introduce you to eLEGS that is worn by Amanda Boxtel that 19 years ago was spinal cord injured, and as a result of that she has not been able to walk for 19 years until now.}'. The color shade indicates the attention weight, and blank area indicates that the attention is masked out.}
\label{fig:attn_vis3}
\end{figure*}

We visualize the proposed segmented attention, the previous uni-directional attention and bi-directional attention in Figure \ref{fig:attn_vis1}, \ref{fig:attn_vis2} and \ref{fig:attn_vis3}, including the speech with various lengths. 

\textbf{Comprehensive} Compared with bi-directional attention, uni-directional attention obviously loses some information from the subsequent speech features \citep{future-guided,laf,iranzo-sanchez-etal-2022-simultaneous}. Segmented attention applies bidirectional attention within a segment and thereby can get a more comprehensive representation. In Figure \ref{fig:attn_vis1}, compared with uni-directional attention, segmented attention is more similar with bi-directional attention in attention distribution.

\textbf{Precise} DiSeg learns the translation-beneficial segmentation through the proposed expected segmented attention (refer to Eq.(\ref{eq:expected segmented attention})), which encourages the model to segment the inputs at the feature $a_{i}$ if $a_{i}$ does not need to pay attention to subsequent features. As shown in Figure \ref{fig:attn_vis12}, \ref{fig:attn_vis22} and \ref{fig:attn_vis32}, DiSeg can learn precise segmentation and ensure the acoustic integrity in each segment. In particular, the segmentation shown in Figure \ref{fig:attn_vis12} and \ref{fig:attn_vis22} almost guarantees that each speech segment corresponds to a word in the transcription. Note that the last segment in Figure \ref{fig:attn_vis12} and \ref{fig:attn_vis22} often corresponds to silence at the end of speech. Since DiSeg learns segmentation without labeled segmentation/alignment data, the proposed segmented attention can be applied to more streaming tasks.
\begin{figure*}[t]
\centering
\subfigure[Case Ted\_1337\_63 on MuST-C En$\rightarrow$De. We show the results of DiSeg and Wait-k when both lagging 3 segments, where the latency (AL) of DiSeg and Wait-k are 733$ms$ and 1060$ms$ respectively.]{
\includegraphics[width=6.1in]{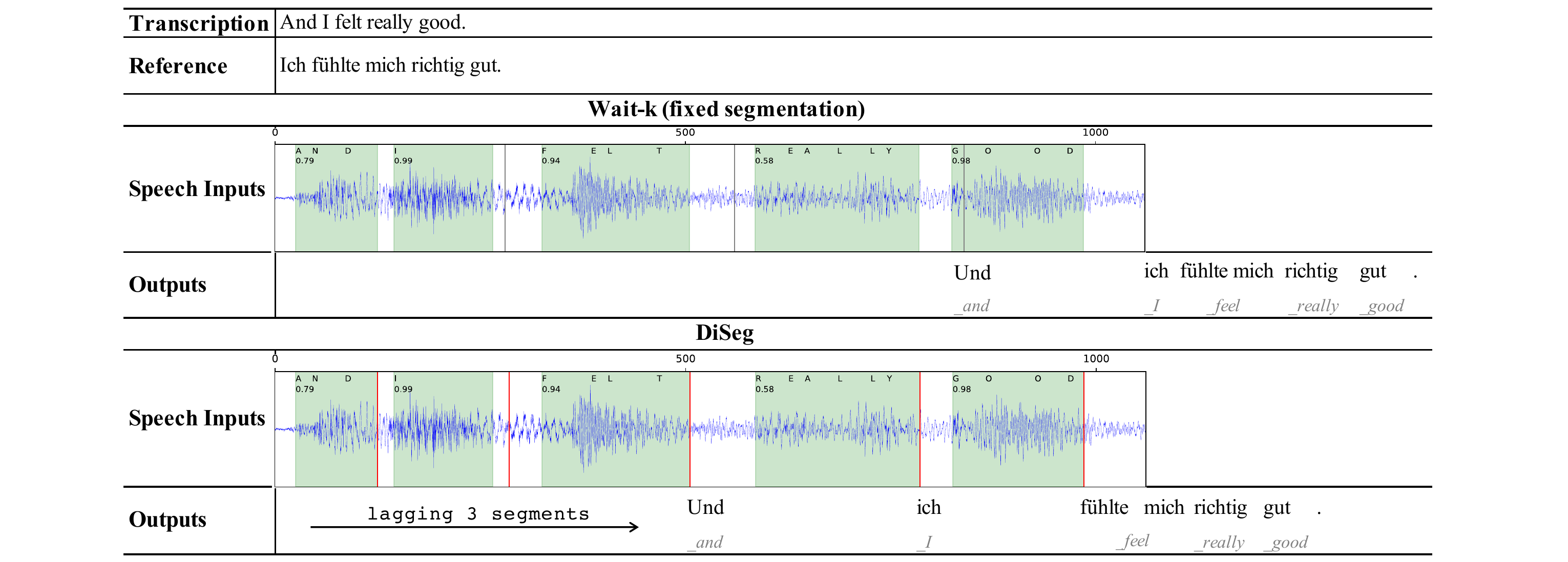}\label{fig:case_study1}
}
\subfigure[Case Ted\_1337\_18 on MuST-C En$\rightarrow$De. We show the results of DiSeg and Wait-k under the same latency, i.e., AL$\approx$750$ms$.]{
\includegraphics[width=6.1in]{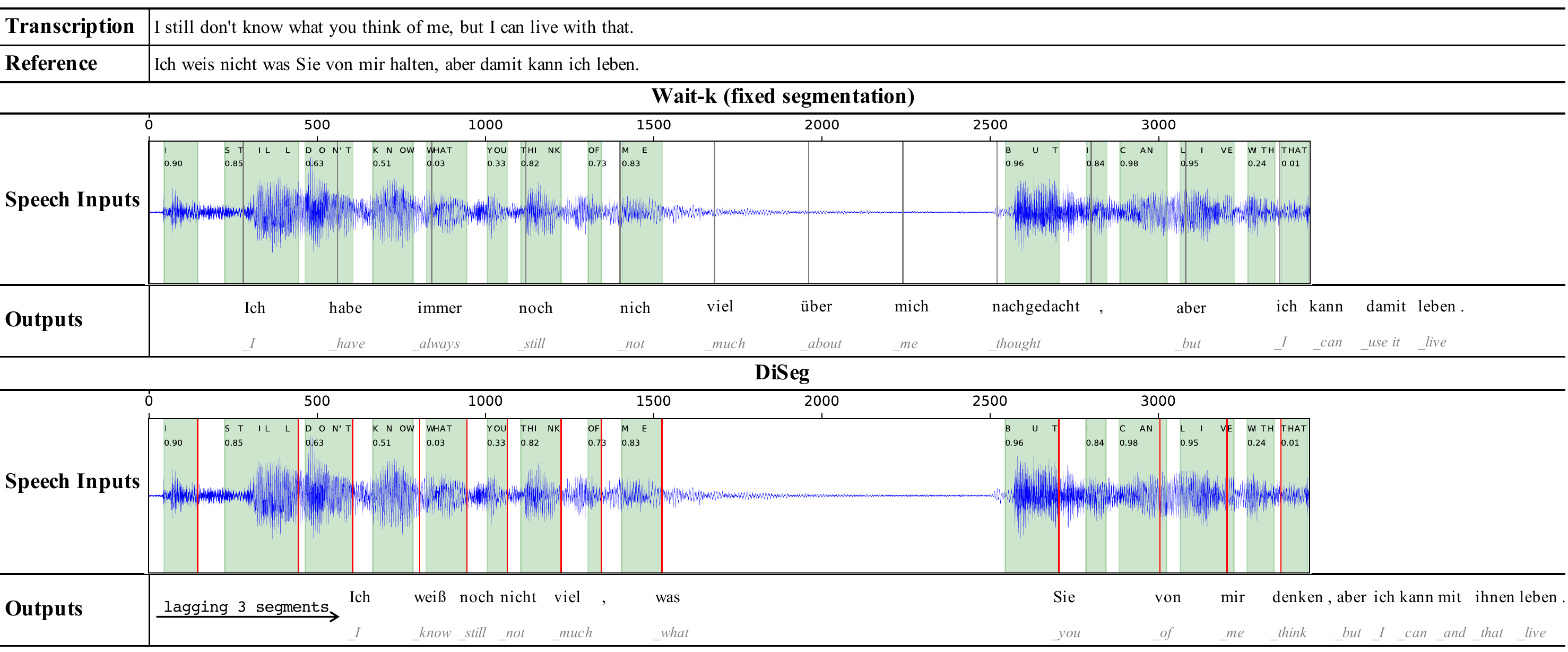}\label{fig:case_study2}
}
\caption{Case study of DiSeg. The horizontal direction indicates when the model outputs the target word with the streaming speech inputs. Red lines indicate where DiSeg decides to segment the speech inputs, and gray lines indicate the equal-length segmentation of 280$ms$. The green areas indicate the alignments between the transcription and the speech generated by an offline alignment tool.}
\label{fig:case_study}
\end{figure*}

\textbf{Concentrate} The issue of attention dispersion caused by long speech is one of the major challenges for speech modeling \citep{YANG2020121,liang2021transformer,valentini2021detection,9054148}. As shown in Figure \ref{fig:attn_vis31} and \ref{fig:attn_vis33}, both uni-directional and bi-directional attention tends to become scattered when dealing with long speech, and each feature can only get a very small amount of attention weight (e.g., the maximum attention weight in Figure \ref{fig:attn_vis33} is $0.01$.), which affects the modeling capability of the attention mechanism \citep{vig-belinkov-2019-analyzing,ding-etal-2019-saliency,valentini2021detection}. Segmented attention applies bi-directional attention within a segment and uni-directional attention between segments, which naturally introduces locality to attention, thereby effectively mitigating the issue of attention dispersion \citep{luong-etal-2015-effective,yang-etal-2018-modeling,liang2021transformer,zhang-feng-2021-modeling-concentrated,9054148}. Specifically, as shown in Figure \ref{fig:attn_vis22}, segmented attention can be concentrated in the segment and pay more attention to the surrounding features. As shown in Figure \ref{fig:attn_vis32}, although the sequence of speech features is extremely long, segmented attention also can focus on the features in each segment (e.g., a clear attention distribution can be found inside each segment in Figure \ref{fig:attn_vis32}, and the maximum attention weight is $0.47$.). Therefore, segmented attention provides a solution to enhance locality in long speech modeling.

\subsection{Case Study}

We visualize the simultaneous translation process of DiSeg on simple and hard cases in Figure \ref{fig:case_study}. Horizontally, the position of the word in outputs is the moment when it is translated, corresponding to the speech inputs. Red lines indicate where DiSeg decides to segment the speech inputs, and gray lines indicate the fixed segmentation of 280$ms$. For clarity, we use an external alignment tool \texttt{Forced-Alignment}\footnote{A CTC-based alignment tool based on the full speech and ground-truth transcription, the tutorial of which can be found at  \url{https://pytorch.org/audio/main/tutorials/forced_alignment_tutorial.html}} \citep{10.1007/978-3-030-60276-5_27} to align the transcription with the speech, where the green area is the speech interval corresponding to the transcription marked by the tool. Note that the alignment provided by external tools is only a rough reference, not necessarily absolutely accurate, and the value in the marked interval represents the probability of alignment.

For the simple case in Figure \ref{fig:case_study1}, where the speech is short and the correspondence between reference and speech (transcription) is much monotonic, DiSeg can basically accurately segment the speech inputs and achieve high-quality translation. In particular, when both lagging 3 segments, DiSeg achieves much lower latency than Wait-k due to more precise segmentation.

For the hard case in Figure \ref{fig:case_study2}, where the speech is much longer and contains a long silence, DiSeg can also precisely segment the speech inputs. Besides, there is an obvious word order difference \citep{gma} between reference and speech (transcription) in this case, which is more challenging for SimulST \citep{ma-etal-2019-stacl}. Since the fixed segmentation cannot adjust, Wait-k misses translating `\textit{know}'. DiSeg can dynamically adjust the segmentation, and thereby decides to segment and translate `\textit{weiß}' after receiving `\textit{know}' in the speech. Owing to precise segmentation, DiSeg can achieve better translation quality under the same latency.

\section{Numerical Results}
\label{app:numerical}

\subsection{Metrics}
For latency, besides Average Lagging (AL) \citep{ma-etal-2019-stacl}, we additionally use Consecutive Wait (CW) \citep{gu-etal-2017-learning}, Average Proportion (AP) \citep{Cho2016} and Differentiable Average Lagging (DAL) \citep{Arivazhagan2019} to evaluate the latency of DiSeg. Assuming that DiSeg translates $y_{t}$ at the moment $\mathcal{T}\!\left ( y_{t} \right )$, the calculations of latency metrics are as follows.

\textbf{Consecutive Wait} \citep{gu-etal-2017-learning} CW evaluates the average waiting duration between two adjacent outputs, calculated as:
\begin{gather}
    \mathrm{CW}=\frac{\sum_{t=1}^{\left | \mathbf{y} \right |} (\mathcal{T}\!\left ( y_{t} \right )-\mathcal{T}\!\left ( y_{t-1} \right ))}{\sum_{t=1}^{\left | \mathbf{y} \right |}\mathbbm{1}_{\mathcal{T}\!\left ( y_{t} \right )-\mathcal{T}\!\left ( y_{t-1} \right )>0}},
\end{gather}
where $\mathbbm{1}_{\mathcal{T}\!\left ( y_{t} \right )-\mathcal{T}\!\left ( y_{t-1} \right )>0}$ counts the number of $\mathcal{T}\!\left ( y_{t} \right )-\mathcal{T}\!\left ( y_{t-1} \right )>0$.

\textbf{Average Proportion} \citep{Cho2016} AP evaluates the average proportion between $\mathcal{T}\!\left ( y_{t} \right )$ and the total duration $T$ of the complete source speech, calculated as:
\begin{gather}
    \mathrm{AP}=\frac{1}{\left | \mathbf{y} \right |}\sum_{t=1}^{\left | \mathbf{y} \right |} \frac{\mathcal{T}\!\left ( y_{t} \right )}{T}.
\end{gather}

\textbf{Average Lagging} \citep{ma-etal-2019-stacl,ma-etal-2020-simulmt} AL evaluates the average duration that target outputs lag behind the speech inputs, is calculated as:
\begin{align}
    \mathrm{AL}=&\; \frac{1}{\tau }\sum_{t=1}^{\tau}\mathcal{T}\!\left ( y_{t} \right )-\frac{t-1}{\left | \mathbf{y} \right | /\; T},\\
    \text{where} \;\;\tau =&\; \underset{t}{\mathrm{argmin}}\left ( \mathcal{T}\!\left ( y_{t} \right )= T\right ).
\end{align}

\textbf{Differentiable Average Lagging} \citep{Arivazhagan2019} DAL is a differentiable version of average lagging, calculated as:
\begin{gather}
\mathrm{DAL}=\frac{1}{\left | \mathbf{y} \right | }\sum\limits_{t=1}^{\left | \mathbf{y} \right |}\mathcal{T}^{'}\!\left ( y_{t} \right )-\frac{t-1}{\left | \mathbf{y} \right | / \;T},
\end{gather}
where
\begin{gather}
\mathcal{T}^{'}\!\!\left ( y_{t} \right )\!= \!\left\{\begin{matrix}
\mathcal{T}\!\left ( y_{t} \right ) & t=1\\ 
 \!\mathrm{max}\!\left (\!\mathcal{T}\!\left ( y_{t} \right ),\mathcal{T}^{'}\!\!\left ( y_{t-1} \right )\!+\! \frac{T}{\left | \mathbf{y} \right |} \right )& t>1
\end{matrix}\right..
\end{gather}

For translation quality, in addition to SacreBLEU \citep{post-2018-call}, we also provide TER \citep{snover-etal-2006-ter}, chrF \citep{popovic-2015-chrf} and chrF++ \citep{popovic-2017-chrf} score of DiSeg.

\subsection{Numerical Results}

The numerical results of DiSeg with more metrics are reported in Table \ref{tab:res_en_de} and Table \ref{tab:res_en_es}.

\begin{table*}[]
\centering
\begin{tabular}{ccccccccc} \toprule
\multicolumn{9}{c}{\textbf{En}$\rightarrow$\textbf{De}}                                           \\ \midrule\midrule
$k$  & \textbf{CW}   & \textbf{AP}   & \textbf{AL}   & \textbf{DAL}  & \textbf{BLEU} & \textbf{TER}   & \textbf{chrF}  & \textbf{chrF++} \\\cmidrule(lr){1-1}\cmidrule(lr){2-5}\cmidrule(lr){6-9}
1  & 462  & 0.67 & 1102 & 1518 & 18.85     & 73.13 & 44.29 & 42.31  \\
3  & 553  & 0.76 & 1514 & 1967 & 20.74     & 69.95 & 49.34 & 47.09  \\
5  & 666  & 0.82 & 1928 & 2338 & 22.11     & 66.90 & 50.13 & 47.94  \\
7  & 850  & 0.86 & 2370 & 2732 & 22.98     & 65.42 & 50.36 & 48.23  \\
9  & 1084 & 0.90 & 2785 & 3115 & 23.01     & 65.48 & 50.24 & 48.13  \\
11 & 1354 & 0.92 & 3168 & 3464 & 23.13     & 65.04 & 50.42 & 48.31  \\
13 & 1632 & 0.94 & 3575 & 3846 & 23.05     & 64.85  & 50.53  & 48.41   \\
15 & 1935 & 0.96 & 3801 & 4040 & 23.12     & 64.92 & 50.47 & 48.36 \\ \bottomrule
\end{tabular}
\caption{Numerical results of DiSeg on MuST-C En$\rightarrow$De.}
\label{tab:res_en_de}
\end{table*}

\begin{table*}[]
\centering
\begin{tabular}{ccccccccc}\toprule
\multicolumn{9}{c}{\textbf{En}$\rightarrow$\textbf{Es}}                                                                                                      \\\midrule\midrule
$k$  & \textbf{CW} & \textbf{AP} & \textbf{AL} & \textbf{DAL} & \textbf{BLEU} & \textbf{TER} & \textbf{chrF} & \textbf{chrF++} \\\cmidrule(lr){1-1}\cmidrule(lr){2-5}\cmidrule(lr){6-9}
1  & 530         & 0.67        & 1144        & 1625         & 22.03         & 71.34        & 45.69          & 43.82            \\
3  & 563         & 0.76        & 1504        & 2107         & 24.49         & 66.63        & 53.09          & 50.85            \\
5  & 632         & 0.81        & 1810        & 2364         & 26.58         & 63.35        & 54.55          & 52.39            \\
7  & 788         & 0.85        & 2249        & 2764         & 27.81         & 61.87        & 55.28          & 53.16            \\
9  & 1010        & 0.89        & 2694        & 3164         & 28.33         & 60.98        & 55.51          & 53.40            \\
11 & 1257        & 0.92        & 3108        & 3530         & 28.59         & 60.63        & 55.64          & 53.55            \\
13 & 1534        & 0.94        & 3479        & 3855         & 28.72         & 60.49        & 55.61          & 53.53            \\
15 & 1835        & 0.95        & 3819        & 4160         & 28.92         & 60.22        & 55.80          & 53.71 \\ \bottomrule          
\end{tabular}
\caption{Numerical results of DiSeg on MuST-C En$\rightarrow$Es.}
\label{tab:res_en_es}
\end{table*}

\end{document}